\documentclass[10pt,twocolumn,letterpaper]{article}

\usepackage{cvpr}
\usepackage{tikz}
\usetikzlibrary{arrows}
\usepackage{times}
\usepackage{epsfig}
\usepackage{graphicx}
\usepackage{amsmath}
\usepackage{amssymb}
\usepackage[font=small,labelfont=bf]{caption}
\usepackage{subcaption}
\usepackage{hhline}
\usepackage{multirow}
\usepackage{color}

\usepackage[pagebackref=true,breaklinks=true,letterpaper=true,colorlinks,bookmarks=false]{hyperref}

\cvprfinalcopy 

\def\cvprPaperID{631} 

\newcommand{\mboxronny}[1]{%
    \textit{#1}%
}

\ifcvprfinal\fi
\begin{document}

\title{Real-Time Depth Refinement for Specular Objects}

\author{Roy Or - El\textsuperscript{1} 
		\quad Rom Hershkovitz\textsuperscript{1} 
        \quad Aaron Wetzler\textsuperscript{1}\\
        \quad Guy Rosman\textsuperscript{2}  
        \quad Alfred M. Bruckstein\textsuperscript{1} 
        \quad Ron Kimmel\textsuperscript{1}\\
\textsuperscript{1}Technion, 
           Israel Institute of Technology\\
\textsuperscript{2}Computer Science and 
          Artificial Intelligence Lab, MIT\\
{\tt\small royorel@cs.technion.ac.il 
	\tt\small sromh@campus.technion.ac.il 
    \tt\small twerd@cs.technion.ac.il}\\
{\quad \tt\small rosman@csail.mit.edu  
	\quad \tt\small freddy@cs.technion.ac.il
	\quad \tt\small ron@cs.technion.ac.il
}
}

\maketitle

\begin{abstract}
The introduction of consumer RGB-D scanners 
 set off a major boost in 3D computer vision research. 
Yet, the precision of existing depth 
 scanners is not accurate enough to recover fine details of a scanned object. 
While modern shading based depth refinement methods 
 have been proven to work well with Lambertian objects, 
 they break down in the presence of specularities. 
We present a novel shape from shading framework that
 addresses this issue and enhances both diffuse and specular objects' depth profiles. 
We take advantage of the built-in monochromatic IR
 projector and IR images of the RGB-D scanners and
 present a lighting model that accounts for the 
 specular regions in the input image. 
Using this model, we reconstruct the depth map in 
  real-time. 
Both quantitative tests and visual evaluations prove that 
 the proposed method produces state of the art depth
 reconstruction results.        
\end{abstract}
\vspace{-5mm}
\setlength{\belowdisplayskip}{3.1pt} \setlength{\belowdisplayshortskip}{3.1pt}
\setlength{\abovedisplayskip}{3.1pt} \setlength{\abovedisplayshortskip}{3.1pt}
\section{Introduction}
The introduction of commodity RGB-D scanners marked 
 the beginning of a new age for computer vision and
 computer graphics. 
Despite their popularity, such scanners can 
 obtain only the rough geometry of scanned 
 surfaces due to limited depth sensing accuracy. 
One way to mitigate this limitation is to refine the 
 depth output of these scanners using the available 
 RGB and IR images.

A popular approach to surface reconstruction from image
 shading cues is the Shape from Shading (SfS). 
Shape reconstruction from a single image is an 
ill-posed problem since beside the surface 
 geometry, the observed image also depends on 
 properties like  the surface reflectance, the 
 lighting conditions and the viewing direction. 
Incorporating data from depth sensors has proved to 
 be successful in eliminating some of these 
 ambiguities~\cite{hanhigh2013,WZNSIT14,Orel2015CVPR}. 
However, many of these efforts are based on the 
 assumption that the scanned surfaces are fully 
 Lambertian, which limits the variety of objects they
 can be applied to. 
Directly applying such methods to specular objects introduces
 artifacts to the surface in highly specular regions 
 due to the model's inability to account for sudden 
 changes in image intensity.     

Here, we propose a novel real-time framework for depth
 enhancement of non-diffuse surfaces. 
To that end, we use the IR image supplied by the 
 depth scanners. 
 The narrowband nature of the IR projector and IR
 camera provides a controlled lighting environment. 
Unlike previous approaches, we exploit this friendly
 environment to introduce a new lighting model for depth
 refinement that accounts for specular reflections as
 well as multiple albedos. 
To enable our real-time method we directly enhance the
 depth map by using an efficient optimization scheme 
 which avoids the traditional normals refinemet step.     

The paper outline is as follows: 
 Section~\ref{sec:related} reviews previous efforts 
 in the field. 
An overview of the problem is presented in  
 Section~\ref{sec:overview}. 
The new method is introduced in 
 Section~\ref{sec:framework},  with results in
 Section~\ref{sec:results}.  
 Section~\ref{sec:Conclusions} concludes the paper.

\section{Related Efforts}
\label{sec:related}
The classical SfS framework 
 assumes a Lambertian object with constant albedo and 
 a single, distant, lighting source with known direction.
There are several notable methods which solve the   
 classical SfS problem. 
These can be divided into two groups: propagation methods
 and variational ones. 
Both frameworks were extensively researched during 
 the last four decades. 
Representative papers from each school of thought
 are covered  in
 ~\cite{zhang1999shape,durou2008numerical}.

The main practical drawback about classical 
 shape from shading, is that although a diffusive 
 single albedo setup can be easily designed in
 a laboratory, it can be rarely found in more 
 realistic environments. 
As such, modern SfS approaches attempt to reconstruct
 the surface without any assumptions about the scene 
  lighting and/or the object albedos. 
In order to account for the unknown scene conditions,
 these algorithms either use learning techniques to 
  construct priors for the shape and scene parameters, or 
  acquire a rough depth map from a 3D scanner to
  initialize the surface. 

\noindent
\textbf{Learning based methods}. 
Barron and Malik~\cite{BarronTPAMI2015} constructed
 priors from statistical data of multiple images to 
 recover the shape, albedo and illumination of a given 
  input image. 
Kar \etal~\cite{categoryShapesKar15} learn 3D deformable
 models from 2D annotations in order to recover 
  detailed shapes. 
Richter and Roth~\cite{Richter2015CVPR} extract color, 
 textons and silhouette features from a test image to 
 estimate a reflectance map from which patches of objects 
 from a database are rendered and used in a learning 
 framework for regression of surface normals. 
Although these methods produce excellent results, 
 they depend on the quality and size of their training
 data, whereas the proposed axiomatic approach does not 
 require a training stage and is therefore applicable 
 in more general settings. 

\noindent
\textbf{Depth map based methods}. 
Bohme \etal~\cite{bohme2010shading} find a MAP estimate 
 of an enhanced range map by imposing a shading 
  constraint on a probalistic image formation model. 
Yu \etal~\cite{yu2013shading} use mean shift clustering 
 and second order spherical harmonics to estimate the fdepth map
  scene albedos and lighting from a color image. 
These estimations are then combined together to improve 
 the given depth map accuracy. 
Han \etal~\cite{hanhigh2013} propose a quadratic global  
 lighting model along with a spatially varying local 
 lighting model to enhance the quality of the depth 
  profile.
Kadambi \etal~\cite{kadambi2015polarized} fuse normals obtained from polarization cues with rough depth maps to obtain accurate reconstructions. Even though this method can handle specular surfaces, it requires at least three photos to reconstruct the normals and it does not run in real-time. 
Several IR based methods were introduced 
 in~\cite{haque2014high,Choe2014CVPR,Chatterjee2015CVPR,Ti_2015_CVPR}. The authors of~\cite{haque2014high,Chatterjee2015CVPR} 
  suggest a multi-shot photometric stereo approach to 
   reconstruct the object normals. 
Choe \etal~\cite{Choe2014CVPR} refine 3D meshes from 
 Kinect Fusion~\cite{newcombe2011kinectfusion} using IR 
  images captured during the fusion pipeline. 
Although this method can handle uncalibrated lighting, 
 it is niether one-shot nor real-time since a mesh 
 must first be acquired before the refinement process
 begins.
Ti~\etal~\cite{Ti_2015_CVPR} propose a simultaneous time-of flight and  photometric stereo algorithm that utilizes several light sources to produce accurate surface and surface normals. Although this method can be implemented in real time, it requires four shots per frame for reconstruction as opposed to our single shot approach.
More inline with our approach, Wu \etal~\cite{WZNSIT14} 
 use second order spherical harmonics to estimate the 
 global scene lighting, which is then followed by  
 efficient scheme to reconstruct the object. In~\cite{Orel2015CVPR} Or - El \etal introduced a 
 real-time framework for direct depth refinement that 
 handles natural lighting and multiple albedo objects. Both algorithms rely on shading cues from an RGB image 
 taken under uncalibrated illumination with possibly 
 multiple light sources. 
Correctly modeling image specularities under such
 conditions is difficult. 
We propose to overcome the light source ambiguity issue 
 by using the availability of a single IR source with known configuration. 

\section{Overview}
\label{sec:overview}
\begin{figure*}
\centering
\includegraphics[height = 7cm]{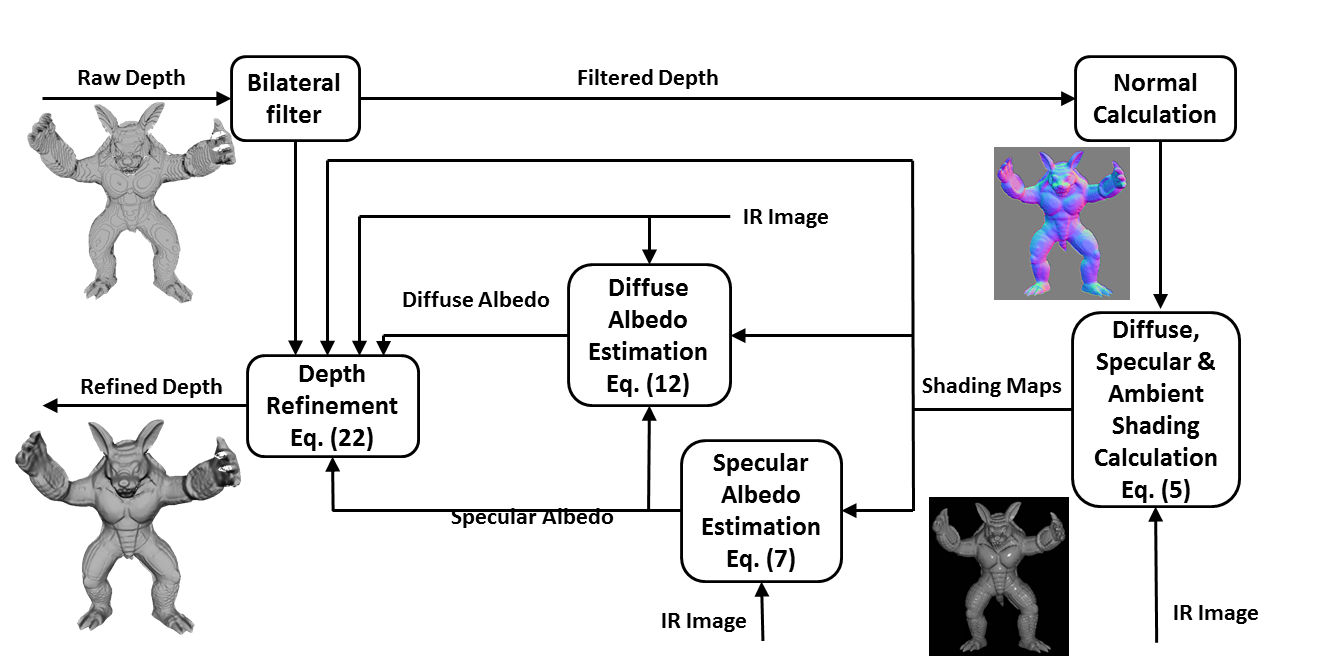}
\vspace{-3mm}
\caption{Algorithm's flowchart}
\label{fig:flowchart}
\vspace{0mm}
\end{figure*}
Shape from Shading (SfS) tries to relate an object's  
 geometry to its image irradiance.  
Like many other inverse problems, SfS is also an 
 ill-posed one because the per-pixel image intensity
 is determined by several elements: the
  surface geometry, its albedo, scene lighting,  
  the camera parameters and the viewing direction.

When using depth maps from RGB-D scanners one could 
 recover the camera parameters and viewing direction, 
  yet, in order to obtain the correct surface, we first 
  need to account for the scene lighting and the surface 
  albedos. 
Failing to do so would cause the algorithm to change 
 the surface geometry and introduce undesired
 deformations.
Using cues from an RGB image under uncalibrated 
 illumination like~\cite{hanhigh2013,WZNSIT14,Orel2015CVPR} 
 requires an  estimation of global lighting parameters. 
Although such estimations work well for diffuse  
 objects, they usually fail when dealing with specular 
 ones and result in a distorted geometry. 
The reason is that specularities are sparse outliers 
 that are not accounted for by classical 
 lighting models. 
Furthermore, trying to use estimated lighting
 directions to model specularities is prone to fail 
 when there are multiple light sources in the scene.    

In our scenario, the main lighting in the IR 
 image comes from the scanner's projector, which 
 can be treated  as a point light source. 
Observe that in this setting, we do not need to 
 estimate a global lighting direction, instead, we 
 use a near light field model to describe the 
  per-pixel lighting direction.  
Subsequently, we can also account for specularities and
 non-uniform albedo map.

In our setting, an initial depth estimation is given  
 by the scanner.
We avoid the process of computing a refined normal 
 field and then fusing depth with normal estimates,
 which is common to SfS methods,
 and solve directly for the depth. 
This eliminates the need to enforce integrability and
 reduces the problem size by half. 
We deal with the non-linear part by calculating a
 first order approximation of the cost functional and
 thereby achieve real-time performance. 

\section{Proposed Framework}
\label{sec:framework}
A novel IR based real-time framework for depth  
 enhancement is proposed. 
The suggested algorithm requires a depth map and an IR
 image as inputs. 
We assume that the IR camera and the depth camera have
 the same intrinsic parameters, as is usually the case 
 with common depth scanners. 
In addition, we also assume that the whole system is 
 calibrated and that the translation vector between the  
  scanner's IR projector and IR camera is known.

Unfortunately, the raw depth map is usually quantized 
 and the surface geometry is highly distorted. 
Therefore, we first smooth the raw depth map and estimate
 the surface normals. 
We then move on to recover the scene lighting using a
 near-field lighting model which explicitly accounts
 for object albedos and specularities.

After we find the scene lighting along with albedo and 
 specular maps, we can directly update the surface 
 geometry by designing a cost functional that relates
 the depth and IR intensity values at each pixel. 
We also show how the reconstruction process can be
 accelerated in order to obtain real-time performance.
Figure~\ref{fig:flowchart} shows a flowchart of the proposed algorithm.

\subsection{Near Field Lighting Model}
\label{subsec:lighting_model}
Using an IR image as an input provides several advantages
 to the reconstruction process. 
Unlike other methods which require alignment between 
 RGB and depth images, in our case, the depth map and IR 
 image are already aligned as they were 
 captured by the same camera. 
Moreover, the narrowband nature of the IR camera means
 that the main light source in the image is the
 scanner's own IR projector whose location relative to
  the camera is known. 
Therefore, we can model the IR projector as a point 
 light source and use a near field lighting model to
  describe the given IR image intensity at each pixel,
\begin{equation}
I = \frac{a\rho_d}{d_p^2}S_{\mboxronny{diff}} + \rho_dS_{\mboxronny{amb}} + \frac{a\rho_s}{d_p^2}S_{\mboxronny{spec}}.
\label{eq:lighting_model}
\end{equation}
Here, $a$ is the projector intensity which is assumed to 
 be constant throughout the image. 
$d_p$ is the distance of the surface point
 from the projector. 
$\rho_d$ and $\rho_s$ are the diffuse and specular 
 albedos. 
$S_{\mboxronny{amb}}$ is the ambient lighting in the scene, which 
 is also assumed to be constant over the image. 
$S_{\mboxronny{diff}}$ is the diffuse shading function of the image
 which is given by the Lambertian reflectance model
\begin{equation}
S_{\mboxronny{diff}} = \vec{N} \cdot \vec{l}_p.
\label{eq:diffuse_shading}
\end{equation}
The specular shading function $S_{\mboxronny{spec}}$ is set 
 according to the Phong reflectance model   
\begin{equation}
S_{\mboxronny{spec}} = \left (\left (2(\vec{l}_p \cdot \vec{N})\vec{N} - \vec{l}_p \right) \cdot \vec{l}_c \right)^{\alpha},
\label{eq:specular_shading}
\end{equation}
 where $\vec{N}$ is the surface normal, $\vec{l}_p, 
  \vec{l}_c$ are the directions from the surface point to 
  the projector and camera respectively and $\alpha$ is 
  the shininess constant which we set to $\alpha = 2$. Figure~\ref{fig:light_model} describes the scene
 lighting model.
For ease of notation, we define
\begin{equation}
\tilde{S}_{\mboxronny{diff}}
   = \frac{a}{d_p^2}S_{\mboxronny{diff}}, \quad   
    \tilde{S}_{\mboxronny{spec}} 
   = \frac{a}{d_p^2}S_{\mboxronny{spec}}.
\end{equation}
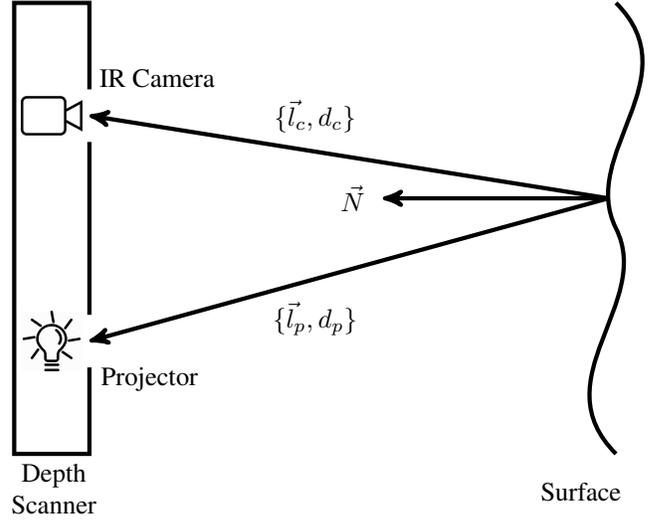
\begin{figure}
	\begin{tikzpicture}
		\begin{scope}[rotate=90, shift={(0,-8)}]
			\draw[ultra thick] (0,0) .. controls (1,1) and (2,-0.5) .. (3,0) .. controls (4,0.5)  and (5,-1) .. (6,0);
     	\end{scope}
        \draw[ultra thick] (1,1.15) -- (1,0) -- (0,0) -- (0,6) -- (1,6) -- (1,4.85);
        \draw[ultra thick] (1,1.85) -- (1,4.15);
        \node (image) at (0.5,1.5) {\includegraphics[height = 0.8cm]{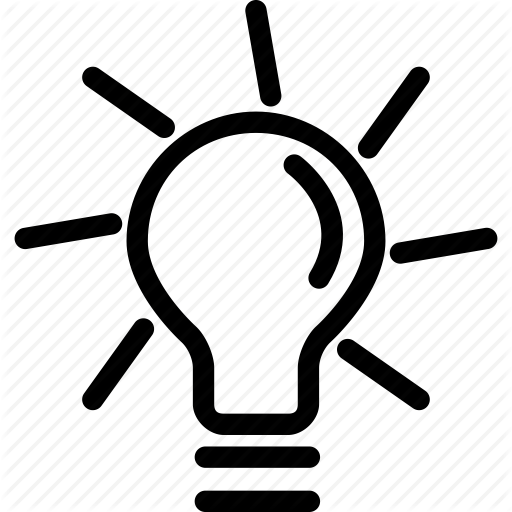}};
        \node (image) at (0.5,4.5) {\includegraphics[height = 0.8cm]{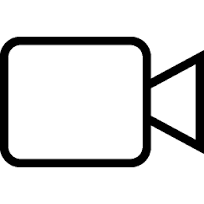}};
        \draw[ultra thick,->,>=stealth'] (7.9,3.4) -- (1,4.5);
        \draw[ultra thick,->,>=stealth'] (7.9,3.4) -- (1,1.5);
        \draw[ultra thick,->,>=stealth'] (7.9,3.4) -- (4.9,3.4);
        \node[text width=1.2cm] (scanner) at (0.35,-0.5) 
        {\begin{tabular}{c} Depth \\ Scanner \end{tabular}};
    	\node[text width=1cm] at (7.5,-0.5) {Surface};
        \node at (4.5,3.4) {$\vec{N}$};
        \node at (4,4.5) {$\{\vec{l}_c,d_c\}$};
        \node at (4,1.8) {$\{\vec{l}_p,d_p\}$};
        \node at (1.8,1) {Projector};
        \node at (1.9,5) {IR Camera};
	\end{tikzpicture}
    \vspace{-8mm}
    \caption{Scene lighting model}
    \label{fig:light_model}
    \vspace{-5.5mm}
\end{figure}

\begin{figure*}
\centering
\begin{subfigure}{0.21\textwidth}
\includegraphics[trim = 20mm 20mm 20mm 20mm, clip=true, height = 4cm]{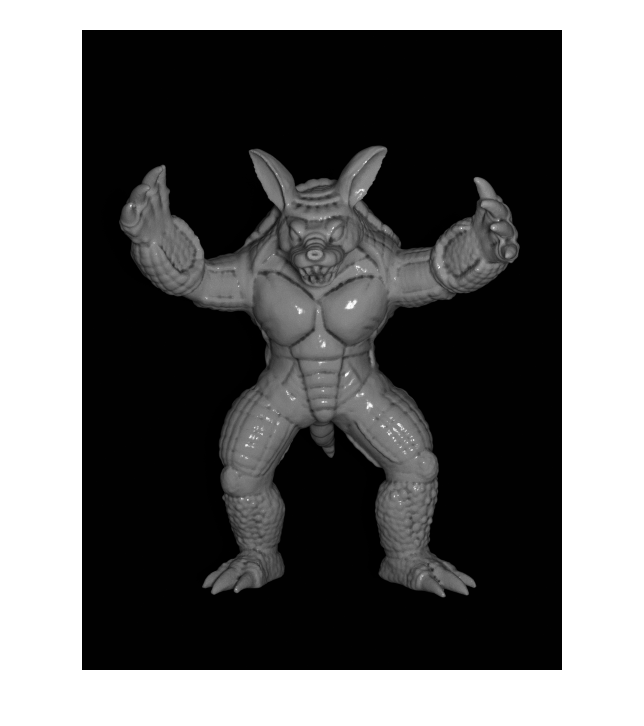}
\caption{}
\end{subfigure}
\begin{subfigure}{0.21\textwidth}
\includegraphics[trim = 20mm 20mm 20mm 20mm, clip=true, height = 4cm]{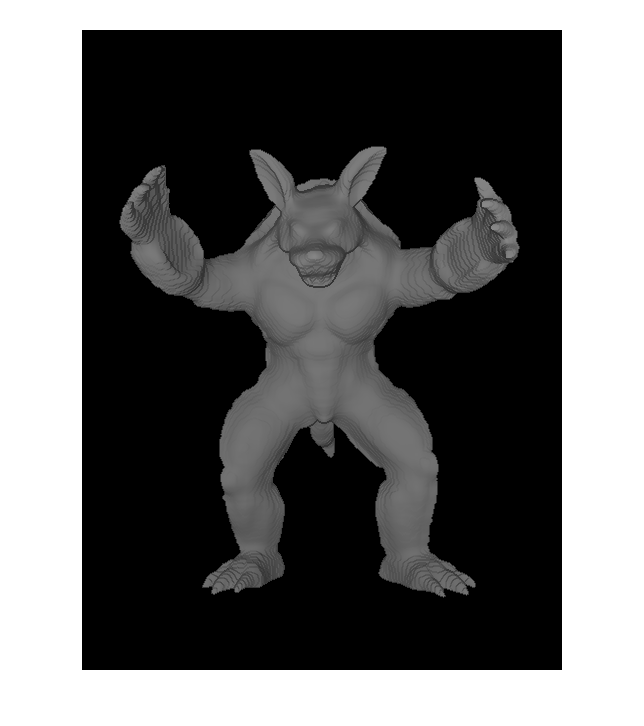}
\caption{}
\end{subfigure}
\begin{subfigure}{0.21\textwidth}
\includegraphics[trim = 20mm 20mm 20mm 20mm, clip=true, height = 4cm]{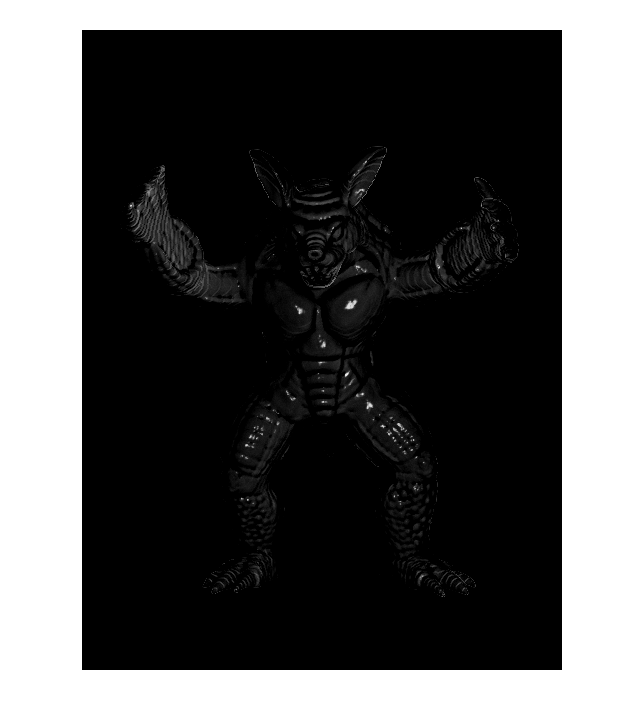}
\caption{}
\end{subfigure}
\begin{subfigure}{0.21\textwidth}
\includegraphics[trim = 20mm 20mm 20mm 20mm, clip=true, height = 4cm]{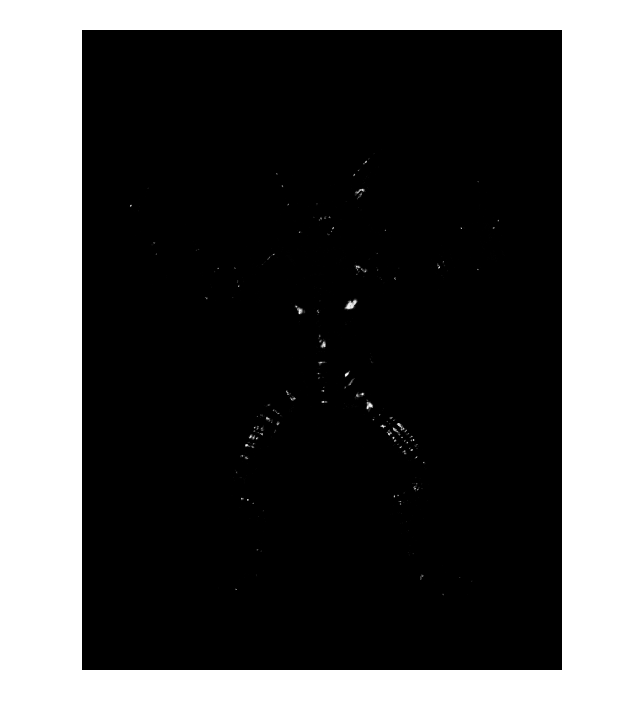}
\caption{}
\end{subfigure}
\vspace{-2mm}
\caption{(a)
 Simulated IR image of the Armadillo mesh. 
 (b) Recovered image of the diffuse and ambient shading 
 $\tilde{S}_{\mboxronny{diff}} + S_{\mboxronny{amb}}$. (c) Residual image for specular albedo estimation $I_{\mboxronny{res}}^s$. (d) Ground 
 Truth specularity map of (a). Note that specularities 
 in (d) are basically the sparse representation of the 
 residual image (c).}
\label{fig:i_res_s}
\vspace{-5mm}
\end{figure*}
The intrinsic camera matrix and the relative location 
 of the projector with respect to camera are known. 
In addition, the initial surface normals can be easily 
 calculated from the given rough surface. 
Therefore, $\vec{l}_c, \vec{l}_p, d_p, S_{\mboxronny{diff}}$ 
 and $S_{\mboxronny{spec}}$ can be found directly whereas 
 $a,S_{\mboxronny{amb}},\rho_d$ and $\rho_s$ need to be recovered. Although we are using a rough depth normal field to 
 compute $\vec{l}_c, \vec{l}_p, d_p, S_{\mboxronny{diff}}$ and 
 $S_{\mboxronny{spec}}$ we still get accurate shading maps since the 
  lighting is not sensitive to minor changes in the depth 
  or normal field as shown 
  in~\cite{basri2003lambertian,ramamoorthi2001efficient}.
Decomposing the IR image into its Lambertian and Specular lighting components along with their respective albedo maps has no unique solution. To achieve accurate results while maintaining real-time performance we choose a greedy approach which first assumes Lambertian lighting and gradually accounts for the lighting model from Eq.~\ref{eq:lighting_model}.
Every pixel in the IR image which has an assigned normal 
 can be used to recover $a$ and $S_{\mboxronny{amb}}$. 
Generally, most of the light reflected back to the
 camera is related to the diffuse component of the
 object whereas highly specular areas usually have a 
 more sparse nature. 
Thus, the specular areas can be treated as outliers 
 in a parameter fitting scheme as they have 
 minimal effect on the outcome. 
This allows us to assume that the object is fully  
 Lambertian (i.e $\rho_d = 1, \rho_s = 0$), which 
 in turn, gives us the following overdetermined linear 
  system for $n$ valid pixels $(n \gg 2)$,
\begin{equation}
	\begin{pmatrix}
		\frac{S_{\mboxronny{diff}}^1}{(d_p^1)^2} & 1 \\
        \vdots & \vdots \\
        \frac{S_{\mboxronny{diff}}^n}{(d_p^n)^2} & 1
	\end{pmatrix}
    \begin{pmatrix}
		a \\
        S_{\mboxronny{amb}}\\
	\end{pmatrix} = 
    \begin{pmatrix}
		I_1\\
        \vdots\\
        I_n\\
	\end{pmatrix}.
\label{eq:ambient_and_projection}
\end{equation}

\vspace{-2.5mm}
\subsubsection{Specular Albedo Map}
\label{subsec:spec_albedo}
The specular shading map is important since it reveals 
 the object areas which are likely to produce specular
 reflections in the IR image. 
Without it, bright diffuse objects can be mistaken for 
 specularities. 
Yet, since $\tilde{S}_{\mboxronny{spec}}$ was calculated as if the 
 object is purely specular, using it by itself will fail
 to correctly represent the specular irradiance, as it  
 would falsely brighten non-specular areas. 
In order to obtain an accurate representation of the
 specularities it is essential to find the specular
 albedo map to attenuate the non-specular areas of
 $\tilde{S}_{\mboxronny{spec}}$.

We now show how we can take advantage of the sparse
 nature of the specularities to recover $\rho_s$ and get 
  the correct specular scene lighting. 
We will define a residual image $I_{\mboxronny{res}}^s$ as being a 
 difference between the original image $I$ and our 
 current diffuse approximation together with 
 the ambient lighting. 
Formally, we write this as
\begin{equation}
 I_{\mboxronny{res}}^s = I - (\tilde{S}_{\mboxronny{diff}} + S_{\mboxronny{amb}}).
\label{eq:specular_reference_image}
\end{equation}
As can be seen in Figure~\ref{fig:i_res_s} (c), the 
 sparse bright areas of $I_{\mboxronny{res}}^s$ are attributable 
  to the true specularities in $I$. 
Specular areas have finite local support, therefore we
 choose to model the residual image $I_{\mboxronny{res}}^s$ as
  $\rho_s\tilde{S}_{\mboxronny{spec}}$ such that $\rho_s$ will be 
  a sparse specular albedo map. 
This will yield an image that contains just the bright
 areas of $I_{\mboxronny{res}}^s$. 
In addition, in order to preserve the smooth nature of
 specularities we add a smoothness term that minimizes
  the L1 Total-Variation of $\rho_s$. 
To summarize, the energy minimization problem to 
 estimate $\rho_s$ can be written as 
\begin{equation}
\underset{\rho_s}{\operatorname{min}}\ \lambda_1^s\|\rho_s\tilde{S}_{\mboxronny{spec}} -I_{\mboxronny{res}}^s\|_2^2 + \lambda_2^s\|\rho_s\|_1 + \lambda_3^s\|\nabla\rho_s\|_1,
\label{eq:specular_Albedo minimization}
\end{equation}
where $\lambda_1^s, \lambda_2^s, \lambda_3^s$ are  
 weighting terms for the fidelity, sparsity and  
 smoothness terms, respectively. 
To minimize the cost functional, we use a variation of the Augmented Lagrangian method suggested  
 in~\cite{Wu:2010} where we substitute the frequency
 domain solution with a Gauss-Seidel scheme on the GPU. 
We refer the reader to the above paper for additional
 details on the optimization procedure.

\subsubsection{Recovering the Diffuse Albedo}
As was the case with specular shading, the diffuse 
 shading map alone does not sufficiently explain the 
 diffuse lighting. 
This is due to the fact that the diffuse shading is
 calculated as if there was only a single object with 
  uniform albedo. 
In reality however, most objects are composed of
 multiple different materials with different reflectance 
  properties that need to be accounted for. 

Using the estimated specular lighting from
 section~\ref{subsec:spec_albedo} we can now compute a 
  residual image between the original image $I$ and the 
  specular scene lighting which we write as
\begin{equation}
I_{res}^d = I - \rho_s\tilde{S}_{\mboxronny{spec}}.
\label{eq:diffuse_reference_image}
\end{equation}
$I_{\mboxronny{res}}^d$ should now contain only the diffuse
 and ambient irradiance of the original image $I$. This can be used 
 in a data fidelity term for a cost functional designed 
 to find the diffuse albedo map $\rho_d$. 

We also wish to preserve the piecewise-smoothness of the 
 diffuse albedo map. 
Otherwise, geometry distortions will be mistaken for 
 albedos and we will not be able to recover the correct  
 surface. 
 The IR image and the rough depth map provide us several 
  cues that will help us to enforce piecewise 
   smoothness. 
Sharp changes in the intensity of the IR image imply a 
 change in the material reflectance. 
Moreover, depth discontinuities can also signal 
 possible changes in the albedo.

We now wish to fuse the cues from the initial depth 
 profile and the IR image together with the piecewise-smooth albedo requirement. 
Past papers~\cite{hanhigh2013,Orel2015CVPR} have used 
 bilateral smoothing. 
Here, instead, we base our scheme on the geomtric Beltrami framework such as in~\cite{sochen1998general,roussos2010tensor,Wetzler11} 
  which has the advantage of promoting alignment of the embedding 
  space channels. 
Let, 
\begin{equation}
\mathcal{M}(x,y) = \{x,y,\beta_II_{\mboxronny{res}}^d(x,y),\beta_zz(x,y),\beta_{\rho}\rho_d(x,y)\}
\label{eq:manifold_definition}
\end{equation}
 be a two dimensional manifold embedded in a $5D$ space 
  with the metric
\begin{equation}
G = \begin{pmatrix}
		\langle\mathcal{M}_x,\mathcal{M}_x\rangle & \langle\mathcal{M}_x,\mathcal{M}_y\rangle\\
        \langle\mathcal{M}_x,\mathcal{M}_y\rangle & \langle\mathcal{M}_y,\mathcal{M}_y\rangle
	\end{pmatrix}.
\label{eq:metric_definition}
\end{equation}
The gradient of $\rho_d$ with respect to the $5D$
 manifold is
\begin{equation}
\nabla_G\rho_d = G^{-1}\cdot\nabla\rho_d,
\label{eq:manifold_gradient_definition}
\end{equation}
By choosing large enough values of $\beta_I,\beta_z$ and 
 $\beta_{\rho}$ and minimizing the L1 Total-Variation of 
 $\rho_d$ with respect to the manifold metric, 
  we basically perform selective smoothing according 
  to the  ``feature'' space $(I_{\mboxronny{res}}^d,z,\rho_d)$. For instance, if $\beta_I \gg \beta_z,\beta_{\rho},1$,
  the manifold gradient would get small values when
  sharp edges are present in $I_{\mboxronny{res}}^d$ since $G^{-1}$
  would decrease the weight of the gradient at such
  locations.

To conclude, the minimization problem we should solve in
 order to find the diffuse albedo map is
\begin{equation}
\underset{\rho_d}{\operatorname{min}}\ \lambda_{1}^d\left \|\rho_d\left(\tilde{S}_{\mboxronny{diff}} + S_{\mboxronny{amb}}\right) - I_{\mboxronny{res}}^d\right \|_2^2 + \lambda_{2}^d\|\nabla_G\rho_d\|_1.
\label{eq:diffuse_albedo_minimization}
\end{equation}
Here, $\lambda_1^d, \lambda_2^d$ are weighting terms for 
 the fidelity and piecewise-smooth penalties. 
We can minimize this functional using the Augmented
 Lagrangian method proposed in~\cite{rosman2012group}.
The metric is calculated separately for each pixel,
 therefore, it can be implemented very efficiently on a
 GPU with limited effect on the algorithm's runtime.

\subsection{Surface Reconstruction}
Once we account for the scene lighting, any differences
 between the IR image and the image rendered with our
 lighting model are attributed to geometry errors of 
 the depth profile. 
Usually, shading based reconstruction algorithms opt to
 use the dual stage process of finding the correct
 surface normals and then integrating them in order to
 obtain the refined depth. 
Although this approach is widely used, it has some
 significant shortcomings. 
Calculating the normal field is an ill-posed problem
 with $2n$ unknowns if $n$ is the number of pixels. 
The abundance of variables can result in distorted
 surfaces that are tilted away from the camera. 
In addition, since the normal field is an implicit
 surface representation, further regularization such as
 the integrability constraint is needed to ensure that
 the resulting normals would represent a valid surface.
This additional energy minimization functional can
 impact the performance of the algorithm.

Instead, we use the strategy suggested
 in~\cite{Orel2015CVPR,WZNSIT14} and take advantage of
 the rough depth profile acquired by the scanner. 
Using the explicit depth values forces the surface to
 move only in the direction of the camera rays, avoids
 unwanted distortions, eliminates the need to use an
 integrability constraint and saves computation time and
 memory by reducing the number of variables.

In order to directly refine the surface, we relate the
 depth values to the image intensity through the surface
 normals. 
Assuming that the perspective camera intrinsic  
 parameters are known, the $3D$ position $P(i,j)$ of 
 each pixel is given by
\begin{equation}
P\left(z(i,j)\right) = \left(\frac{j-c_x}{f_x}z(i,j),\frac{i-c_y}{f_y}z(i,j),z(i,j)\right)^T,
\label{3d_points}
\end{equation}
where $f_x,f_y$ are the focal lengths of the camera and
 $(c_x,c_y)$ is the camera's principal point. 
The surface normal $\vec{N}$ at each $3D$ point is 
 then calculated by
\begin{equation}
\vec{N}\left(z(i,j)\right) = \frac{P_x \times P_y}{\|P_x \times P_y\|}.
\label{eq:normals}
\end{equation}
We can use Eqs.~\eqref{eq:lighting_model},
 ~\eqref{eq:diffuse_shading} and ~\eqref{eq:normals} to
 write down a depth based shading term written directly
 in terms of $z$,
\begin{equation}
E_{sh}(z) = \left\|\frac{a\rho_d}{d_p^2}(\vec{N}(z)\cdot\vec{l}_p) + \rho_dS_{\mboxronny{amb}} + \rho_s\tilde{S}_{\mboxronny{spec}} - I\right\|_2^2.
\label{eq:shading_term}
\end{equation}
This allows us to refine $z$ by penalizing shading
 mismatch with the original image $I$. 
We also use a fidelity term that penalizes the distance
 from the initial 3D points 
\begin{equation}
\begin{aligned}
&E_f(z) = \|w(z-z_0)\|_2^2,\\ 
&w = \sqrt{1 + \left(\frac{j-c_x}{f_x}\right)^2 + \left(\frac{i-c_y}{f_y}\right)^2},
\end{aligned}
\label{eq:fidelity_term}
\end{equation}
 and a smoothness term that minimizes the second order  
 TV-L1 of the surface
\begin{equation}
E_{sm}(z) = \|Hz\|_1, \quad H = \begin{pmatrix}
						D_{xx}\\
                        D_{yy}
					\end{pmatrix}.
\label{eq:smoothness_term}
\end{equation}
Here, $D_{xx}, D_{yy}$ are the second derivatives of the
 surface.

Combining Eqs.~\eqref{eq:shading_term},~\eqref{eq:fidelity_term} and~\eqref{eq:smoothness_term} into a cost functional results in a non-linear optimization problem
\begin{equation}
\underset{z}{\operatorname{min}}\ \lambda_1^zE_{sh}(z) + \lambda_2^zE_f(z) + \lambda_3^zE_{sm}(z),
\label{eq:non_linear_opt}
\end{equation}
where $\lambda_1^z, \lambda_2^z, \lambda_3^z$ are the
 weights for the shading, fidelity and smoothness terms,
 respectively. 
Although there are several possible methods to solve
 this problem, a fast scheme is required for real-time
 performance. 
To accurately and efficiently refine the surface we base
 our approach on the iterative scheme suggested
  in~\cite{ping1994shape}. 
Rewriting Eq.~\eqref{eq:shading_term} as a function of
 the discrete depth map $z$, and using forward
 derivatives we have     
\begin{equation}
\begin{aligned}
&I_{i,j} - \rho_dS_{\mboxronny{amb}} - \rho_s\tilde{S}_{\mboxronny{spec}} = \frac{a\rho_d}{d_p^2}(\vec{N}(z)\cdot\vec{l}_p)\\ 
       &= f(z_{i,j},z_{i+1,j},z_{i,j+1}).
\end{aligned}
\end{equation}
\begin{table}
    \begin{center}
    \begin{tabular}{|c | c | c | c |}
    \hline
    Model & IR & NL - SH1 & NL - SH2\\    
    \hhline{|=|=|=|=|}
    Armadillo & {\color{red} \textbf{2.018}} & 12.813 & 11.631\\
    \hline
    Dragon & {\color{red} \textbf{3.569}} & 10.422 & 10.660\\
    \hline
    Greek Statue & {\color{red} \textbf{2.960}} & 7.241 & 9.067\\
    \hline
    Stone Lion & {\color{red} \textbf{4.428}} & 7.8294 & 8.640\\
    \hline
    Cheeseburger & {\color{red} \textbf{9.517}} & 17.881 & 19.346\\
    \hline
    Pumpkin & {\color{red} \textbf{10.006}} & 13.716 & 16.088\\
    \hline
    \end{tabular}
    \caption{Quantitative comparison of RMSE of the
     specular lighting estimation in IR and natural 
     lighting scenarios. 
     IR refers to the lighting scenario described in
     Section~\ref{subsec:lighting_model}, NL - SH1/2
     represents a natural lighting scenario with
     first/second order spherical harmonics used to
     recover the diffuse and ambient shading as well as
     $\vec{l}_p$. 
     All values are in gray intensity units $[0,255]$.}
    \vspace{-7mm}
    \label{tab:specular_results}
    \end{center}
\end{table}
At each iteration $k$ we can approximate $f$ using the 
 first order Taylor expansion about
  $(z_{i,j}^{k-1},z_{i+1,j}^{k-1},z_{i,j+1}^{k-1})$,
  such that
\begin{equation}
\begin{aligned}
&I_{i,j}  - \rho_dS_{\mboxronny{amb}} - \rho_s\tilde{S}_{\mboxronny{spec}} 
   = f(z_{i,j}^k,z_{i+1,j}^k,z_{i,j+1}^k)\\
       &\approx f(z_{i,j}^{k-1},z_{i+1,j}^{k-1},z_{i,j+1}^{k-1}) + \frac{\partial f}{\partial z_{i,j}^{k-1}}(z_{i,j}^k - z_{i,j}^{k-1})\\  
       & + \frac{\partial f}{\partial z_{i+1,j}^{k-1}}(z_{i+1,j}^k - z_{i+1,j}^{k-1}) 
       + \frac{\partial f}{\partial z_{i,j+1}^{k-1}}(z_{i,j+1}^k - z_{i,j+1}^{k-1}).
\end{aligned}
\end{equation}
Rearranging terms to isolate terms including $z$ from
 the current iteration, we can define
\begin{equation}
\begin{aligned}
I_{\mboxronny{res}}^{z^k} &= I_{i,j} - \rho_dS_{\mboxronny{amb}} 
   - \rho_s\tilde{S}_{\mboxronny{spec}}\\ 
              &\quad - f(z_{i,j}^{k-1},z_{i+1,j}^{k-1},z_{i,j+1}^{k-1}) + \frac{\partial f}{\partial z_{i,j}^{k-1}}z_{i,j}^{k-1}\\ 
              &\quad + \frac{\partial f}{\partial z_{i+1,j}^{k-1}}z_{i+1,j}^{k-1} + \frac{\partial f}{\partial z_{i,j+1}^{k-1}}z_{i,j+1}^{k-1}
\end{aligned},
\end{equation}
 and therefore minimize 
\begin{equation}
\underset{z^k}{\operatorname{min}}\ \lambda_1^z\|Az^k - I_{\mboxronny{res}}^{z^k}\|_2^2 + \lambda_2^z\|w(z^k-z_0)\|_2^2 + \lambda_3^z\|Hz^k\|_1
\end{equation}
 at each iteration with the Augmented Lagrangian method  
  of~\cite{Wu:2010}. 
Here, $A$ is a matrix that represents the linear
 operations performed on the vector $z^k$. 
Finally, we note that this pipeline was implemented on
 an Intel i7 3.4GHz processor with 16GB of RAM and an
  NVIDIA GeForce GTX650 GPU. 
The runtime for a $640 \times 480$ image is 
 approximately $80$ milliseconds. 
\begin{figure}
\centering
\begin{subfigure}{0.14\textwidth}
\includegraphics[trim = 15mm 6.4mm 15mm 7mm, clip=true, height = 2.8cm]{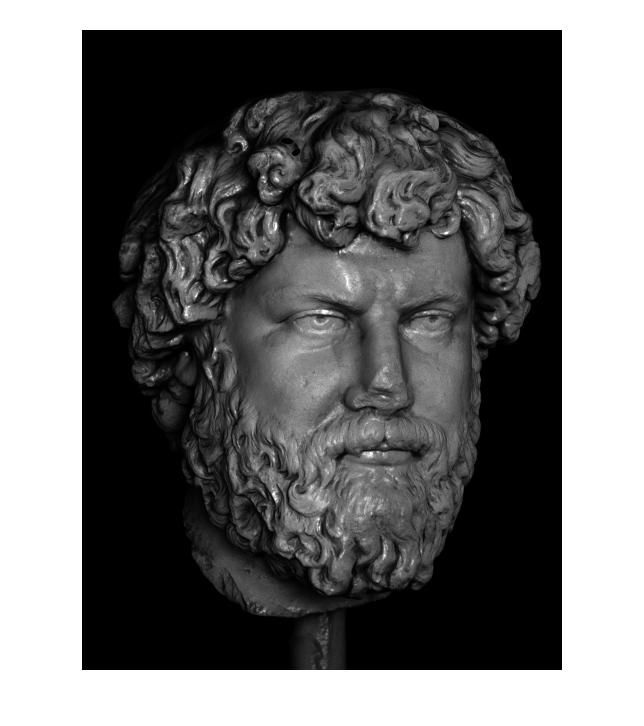}
\caption{}
\end{subfigure}
\hspace{0.01\textwidth}
\begin{subfigure}{0.14\textwidth}
\includegraphics[trim = 15mm 6.4mm 15mm 7mm, clip=true, height = 2.8cm]{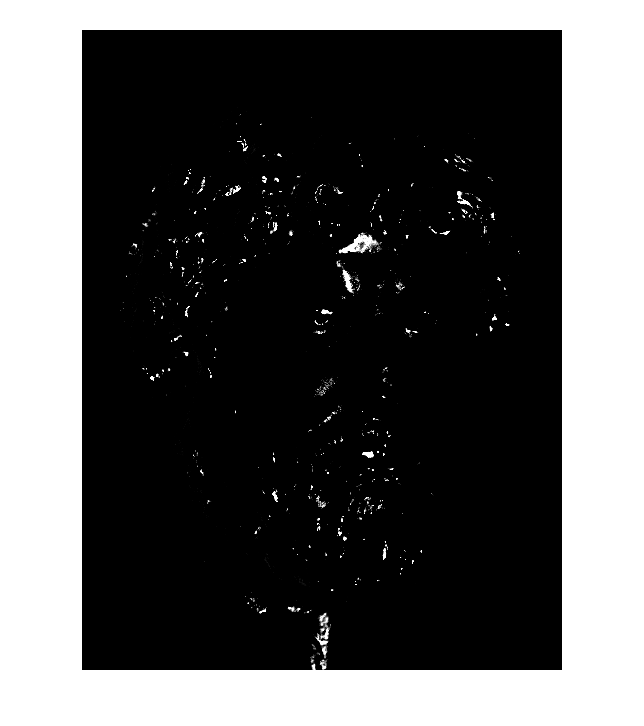}
\caption{}
\end{subfigure}
\hspace{0.01\textwidth}
\begin{subfigure}{0.14\textwidth}
\includegraphics[trim = 15mm 20mm 20mm 18mm, clip=true, height = 2.8cm]{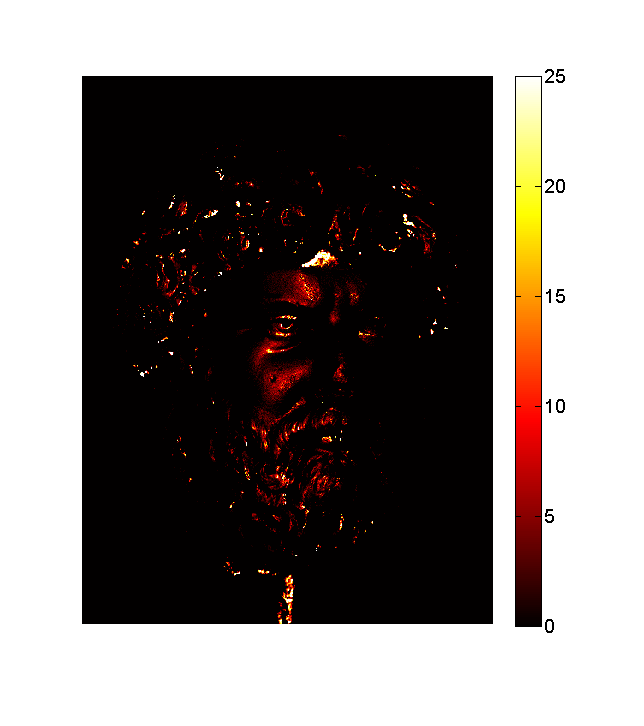}
\caption{}
\end{subfigure}\\
\begin{subfigure}{0.105\textwidth}
\includegraphics[trim = 25mm 6.6mm 15mm 7mm, clip=true, height = 2.5cm]{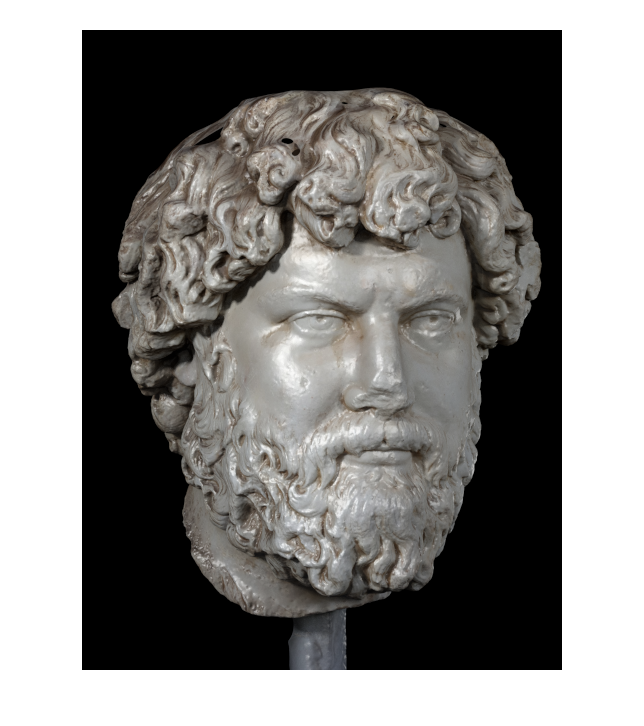}
\caption{}
\end{subfigure}
\hspace{0.00\textwidth}
\begin{subfigure}{0.11\textwidth}
\includegraphics[trim = 15mm 6.6mm 15mm 7mm, clip=true, height = 2.5cm]{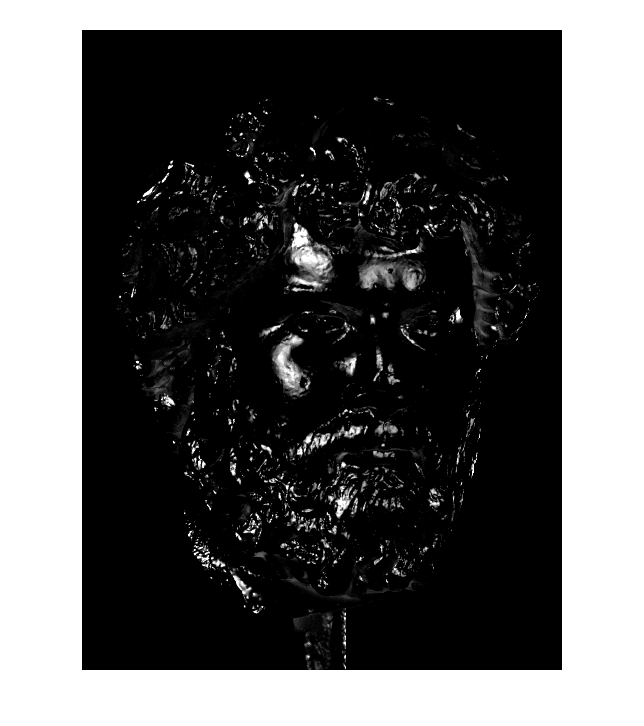}
\caption{}
\end{subfigure}
\hspace{0.005\textwidth}
\begin{subfigure}{0.11\textwidth}
\includegraphics[trim = 25mm 20mm 20mm 18.5mm, clip=true, height = 2.5cm]{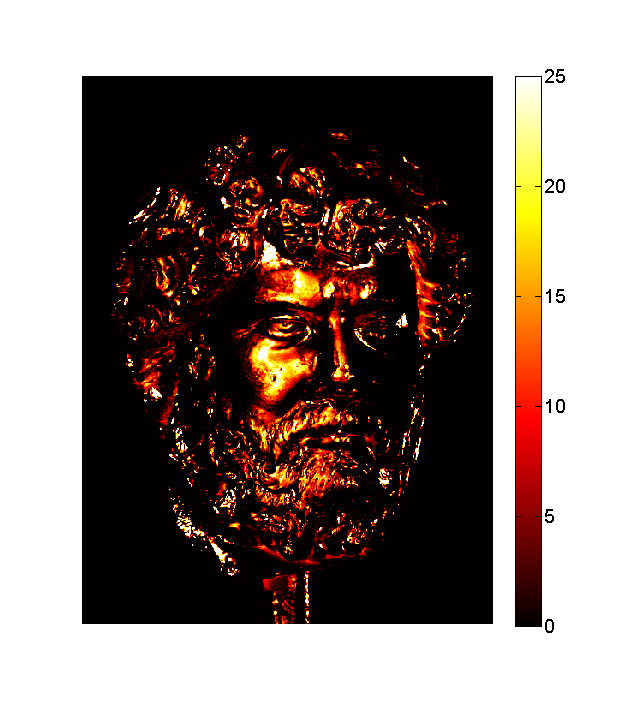}
\caption{}
\end{subfigure}
\hspace{0.005\textwidth}
\begin{subfigure}{0.11\textwidth}
\includegraphics[trim = 25mm 20mm 20mm 18.5mm, clip=true, height = 2.5cm]{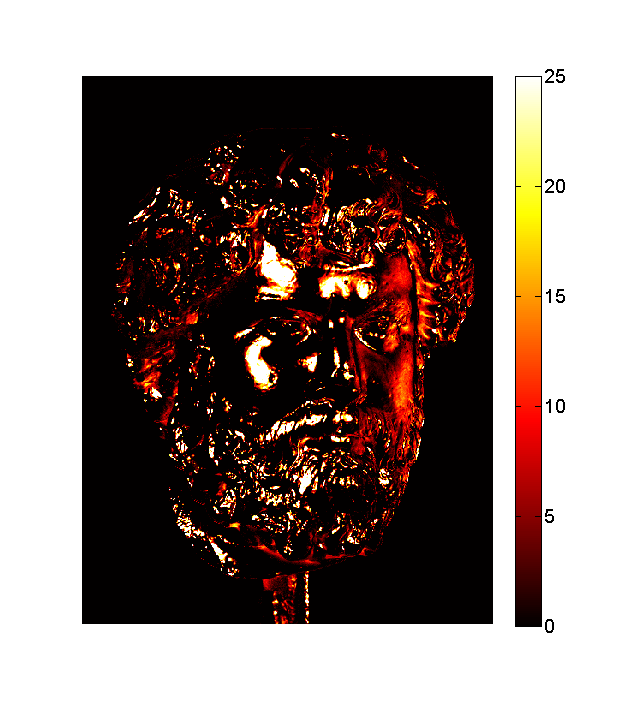}
\caption{}
\end{subfigure}
\vspace{-2mm}
\caption{Greek Statue: 
 (a) Single light source IR image. 
  (b) Ground truth specular irradiance map for (a). 
  (c) Specular irradiance estimation error map. This is the absolute difference map between our predicted specular irradiance and the ground truth.
  (d) Multiple light source natural lighting (NL) image. 
  (e) Specular lighting ground truth of (d). 
  (f,g) Specular irradiance error maps of (d)
  as estimated using first (SH1) and
  second (SH2) order spherical harmonics respectively. 
  Note the reduced errors when using a single known light source (c) as opposed to estimating multiple unknown light sources using spherical harmonics lighting models (f,g).}
\label{fig:specular_test}
\vspace{-5mm}
\end{figure} 
\section{Results}
\label{sec:results}
We preformed several tests in order to evaluate the
 quality and accuracy of the proposed algorithm. 
We show the algorithm's accuracy in recovering the
 specular lighting of the scene and why it is vital to
 use an IR image instead of an RGB image. 
In addition, we demonstrate that the proposed framework
 is state of the art, both visually and qualitatively.

In order to test the specular lighting framework, we 
 took 3D objects from the Stanford  
 $3D$\footnote{http://graphics.stanford.edu/data/3Dscanrep/}, $123D$ Gallery\footnote{http://www.123dapp.com/Gallery/content/all} and 
 Blendswap\footnote{http://www.blendswap.com/} 
 repositories. 
For each model we assigned a mix of diffuse and specular
 shaders and rendered them under an IR lighting scenario
  described in Section~\ref{subsec:lighting_model}
  (single light source) and natural lighting scenarios
  (multiple light sources) using the Cycles renderer in
  Blender. 
To get a ground truth specularity map for each lighting
 scenario, we also captured each model without its
 specular shaders and subtracted the resulting images.

\begin{figure}
\begin{subfigure}{0.14\textwidth}
\includegraphics[trim = 20mm 20mm 20mm 20mm, clip=true, height = 3cm]{graphics/armadillo.png}
\caption{}
\end{subfigure}
\hspace{0.01\textwidth}
\begin{subfigure}{0.14\textwidth}
\includegraphics[trim = 90mm 30mm 85mm 10mm, clip=true, height = 2.8cm]{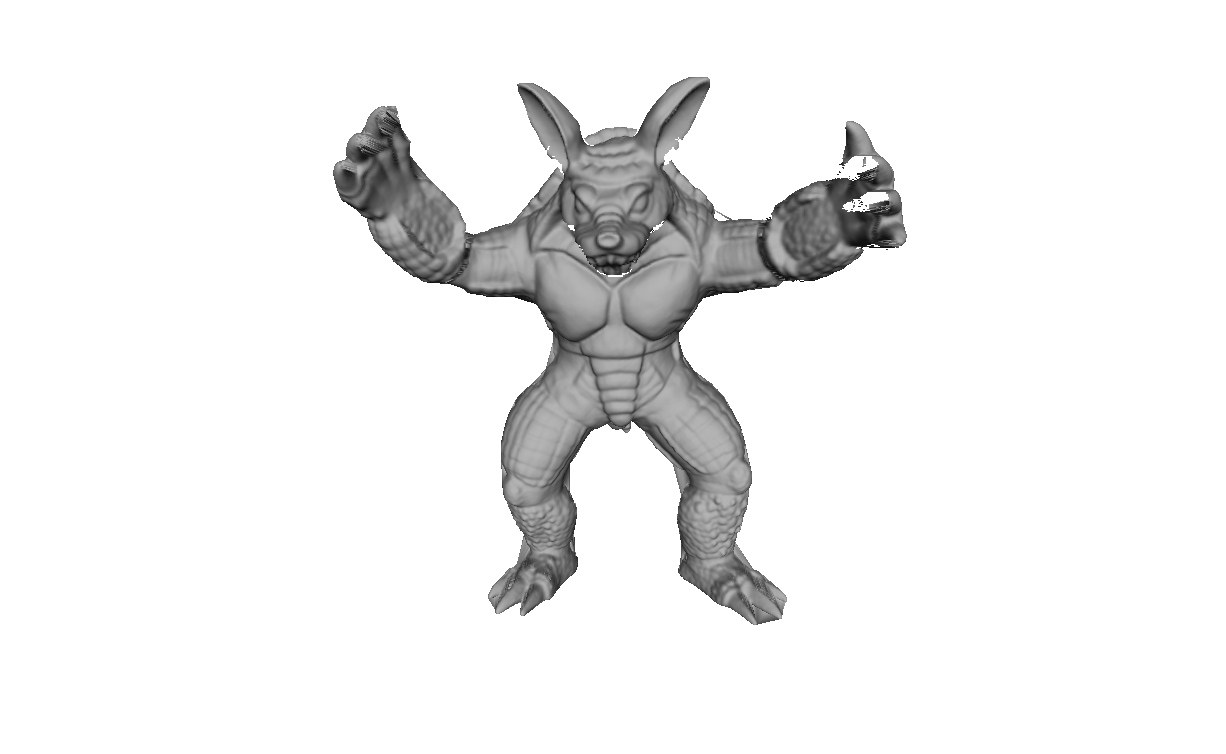}
\caption{}
\end{subfigure}
\hspace{0.01\textwidth}
\begin{subfigure}{0.14\textwidth}
\includegraphics[trim = 90mm 30mm 85mm 10mm, clip=true, height = 2.8cm]{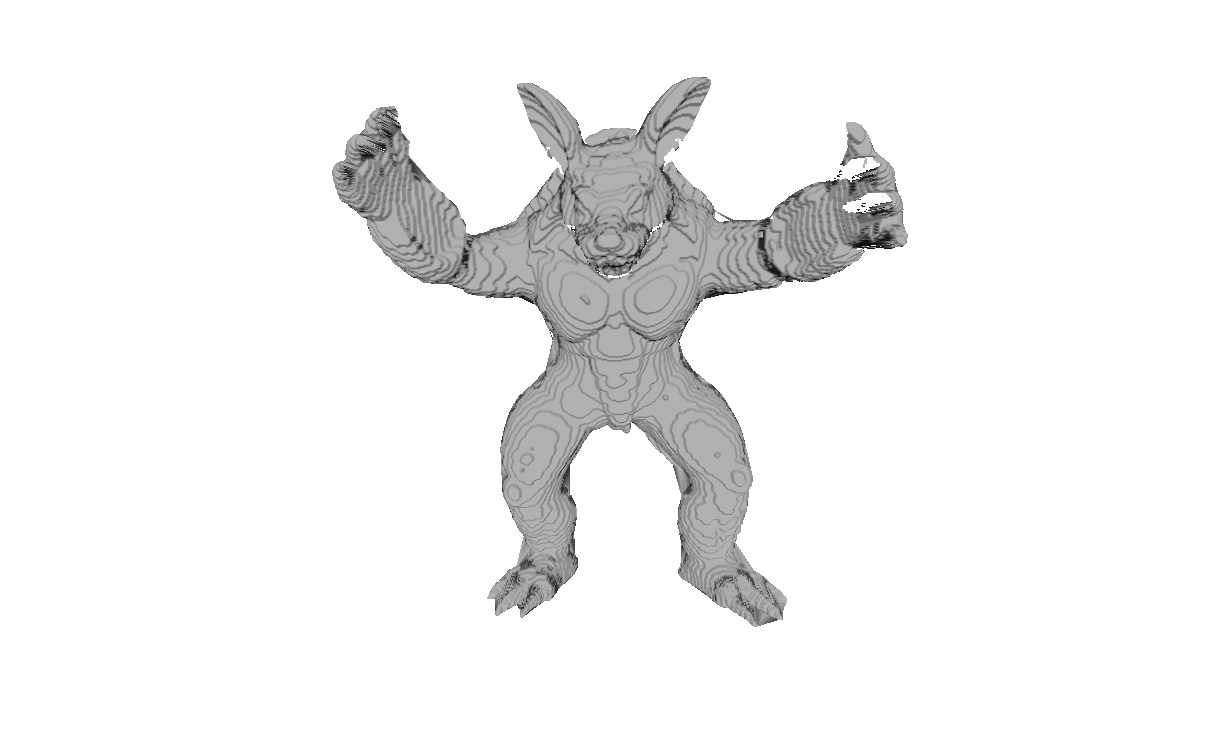}
\caption{}
\end{subfigure}\\
\begin{subfigure}{0.14\textwidth}
\begin{tikzpicture}
\node [anchor=south west,inner sep=0] (image) at (0,0) {\includegraphics[trim = 90mm 30mm 85mm 10mm, clip=true,
   height = 2.8cm]{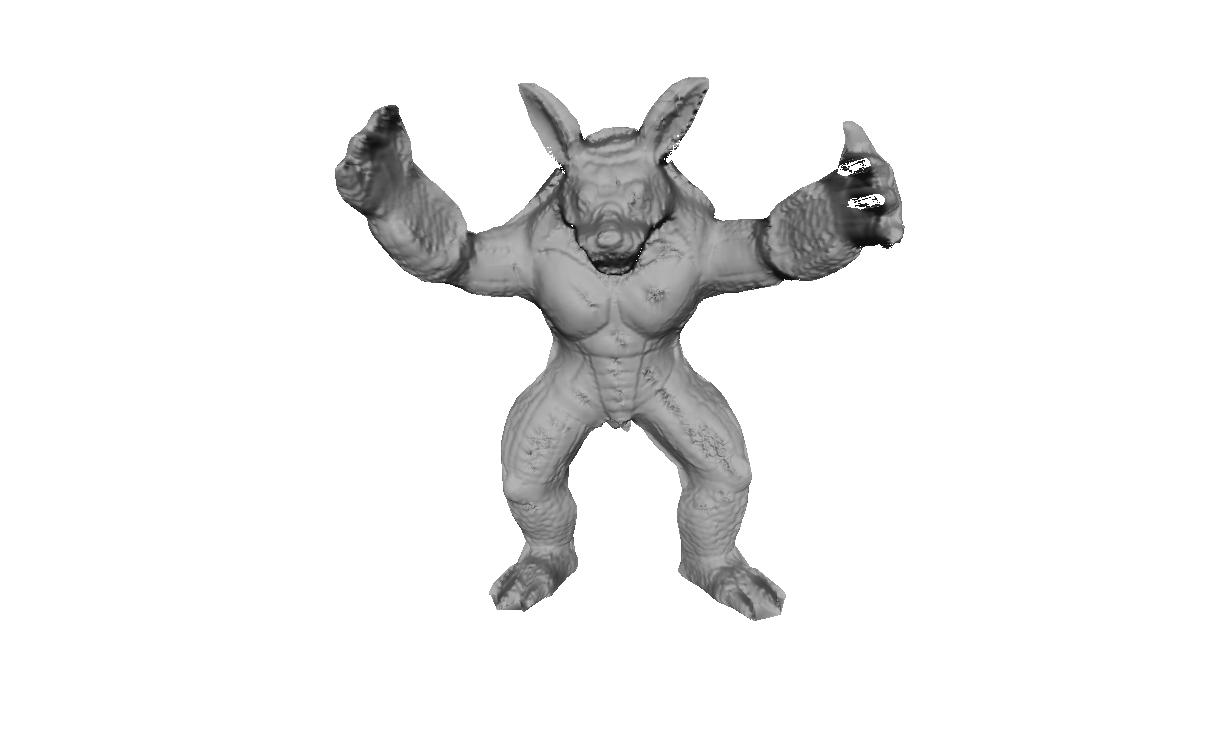}};
\begin{scope}[x={(image.south east)},y={(image.north west)}]
\draw[black,very thick] (0.52,0.52) rectangle (0.62,0.62);
\end{scope}
\end{tikzpicture}
\caption{}
\end{subfigure}
\hspace{0.02\textwidth}
\begin{subfigure}{0.14\textwidth}
\begin{tikzpicture}
\node [anchor=south west,inner sep=0] (image) at (0,0) {\includegraphics[trim = 90mm 30mm 85mm 10mm, clip=true, height = 2.8cm]{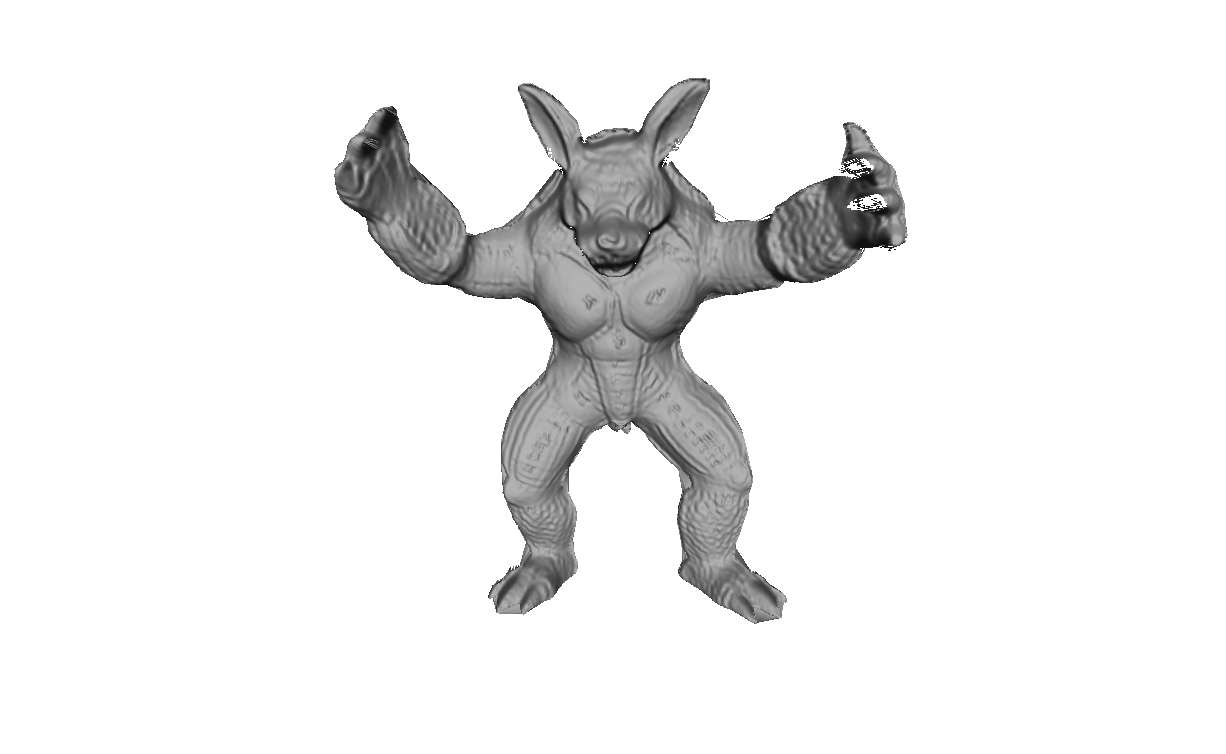}};
\begin{scope}[x={(image.south east)},y={(image.north west)}]
\draw[red,very thick] (0.52,0.52) rectangle (0.62,0.62);
\end{scope}
\end{tikzpicture}
\caption{}
\end{subfigure}
\hspace{0.02\textwidth}
\begin{subfigure}{0.14\textwidth}
\begin{tikzpicture}
\node [anchor=south west,inner sep=0] (image) at (0,0) {\includegraphics[trim = 90mm 30mm 85mm 10mm, clip=true, height = 2.8cm]{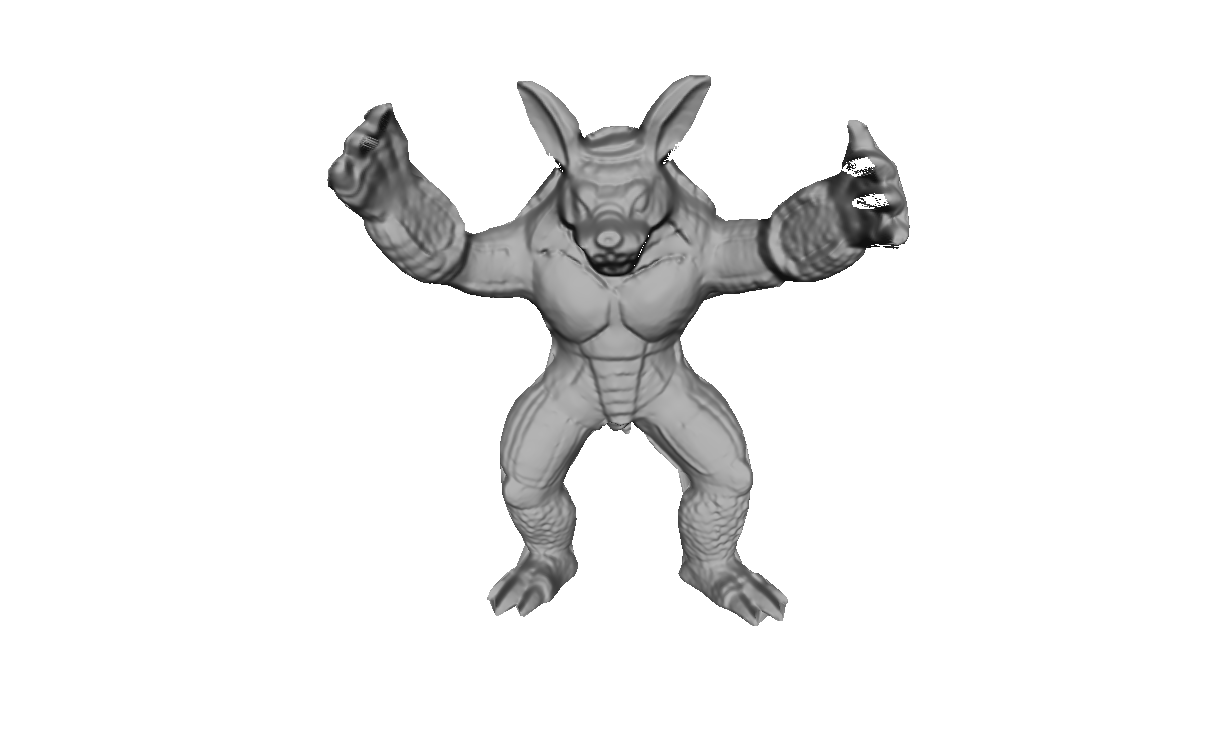}};
\begin{scope}[x={(image.south east)},y={(image.north west)}]
\draw[blue,very thick] (0.52,0.52) rectangle (0.62,0.62);
\end{scope}
\end{tikzpicture}
\caption{}
\end{subfigure}\\
\begin{subfigure}{0.14\textwidth}
\begin{tikzpicture}
\node[anchor=south west,inner sep=0] (image) at (0,0) {\includegraphics[trim = 167mm 115mm 145mm 72mm, clip=true, height = 2.7cm]{graphics/armadillo_wu00.png}};
\begin{scope}[x={(image.south west)},y={(image.north east)}]
\draw[black,very thick] (0.03,0) rectangle (0.97,1);
\end{scope}
\end{tikzpicture}
\caption{}
\end{subfigure}
\hspace{0.01\textwidth}
\begin{subfigure}{0.14\textwidth}
\begin{tikzpicture}
\node[anchor=south west,inner sep=0] (image) at (0,0) {\includegraphics[trim = 167mm 115mm 145mm 72mm, clip=true, height = 2.7cm]{graphics/armadillo_rgbd_fusion00.png}};
\begin{scope}[x={(image.south west)},y={(image.north east)}]
\draw[red,very thick] (0.03,0) rectangle (0.97,1);
\end{scope}
\end{tikzpicture}
\caption{}
\end{subfigure}
\hspace{0.01\textwidth}
\begin{subfigure}{0.14\textwidth}
\begin{tikzpicture}
\node[anchor=south west,inner sep=0] (image) at (0,0) {\includegraphics[trim = 167mm 115mm 145mm 72mm, clip=true, height = 2.7cm]{graphics/armadillo_ours00.png}};
\begin{scope}[x={(image.south west)},y={(image.north east)}]
\draw[blue,very thick] (0.03,0) rectangle (0.97,1);
\end{scope}
\end{tikzpicture}
\caption{}
\end{subfigure}
\vspace{-2mm}
\caption{Results for the simulated Armadillo scene, 
 (a) Input IR image. 
 (b) Ground truth model. 
 (c) Initial Depth. 
 (d)-(f) Reconstructions of Wu \etal, Or - El \etal 
 and our proposed method respectively. 
 (g)-(i) Magnifications of a specular area. 
 Note how our surface is free from distortions in
 specular areas unlike the other methods.}
\label{fig:armadillo_results}
\vspace{-6mm}
\end{figure}
\begin{table*}
    \begin{center}
    \begin{tabular}{| c | c | c | c | c | c | c |}
        \hline
        Model
        \multirow{2}{*} {} &
        \multicolumn{3}{c|} {Median Error (mm)} &
        \multicolumn{3}{c|} {90\textsuperscript{th} \% (mm)} \\
        \cline{2-7}
        & Wu \etal & Or-El \etal & Proposed & Wu \etal & Or-El \etal & Proposed\\
        \hhline{|=|=|=|=|=|=|=|}
        Armadillo & 0.335 & 0.318 & {\color{red}\textbf{0.294}} & 1.005 & 0.821 & {\color{red}\textbf{0.655}}\\
        \hline
        Dragon & 0.337 & 0.344 & {\color{red}\textbf{0.324}} & 0.971 & 0.917 & {\color{red}\textbf{0.870}}\\
        \hline
        Greek Statue & 0.306 & 0.281 & {\color{red}\textbf{0.265}} & 0.988 & 0.806 & {\color{red}\textbf{0.737}}\\
        \hline
        Stone Lion & 0.375 & 0.376 & {\color{red}\textbf{0.355}} & {\color{red}\textbf{0.874}} & 0.966 & 0.949\\
        \hline
        Cheeseburger & 0.191 & 0.186 & {\color{red}\textbf{0.168}} & 0.894 & {\color{red}\textbf{0.756}} & 0.783\\
        \hline
        Pumpkin & 0.299 & 0.272 & {\color{red}\textbf{0.242}} & 0.942 & 0.700 & {\color{red}\textbf{0.671}}\\
        \hline
    \end{tabular}
    \caption{Quantitative comparison of depth accuracy in specular areas. All values are in millimeters.}
    \vspace{-7mm}
    \label{tab:synthetic_results}
    \end{center}
\end{table*}

We tested the accuracy of our model in recovering
 specularities for each lighting setup. 
We used Eqs.~\eqref{eq:diffuse_shading}
 and~\eqref{eq:ambient_and_projection} to get the
 diffuse and ambient shading maps under IR lighting. 
For natural lighting, the diffuse and ambient shading
 were recovered using first and second order spherical
 harmonics in order to have two models for comparison. In both lighting scenarios the surface normals were
 calculated from the ground truth depth map. 
The specular lighting is recovered using
 Eqs.~\eqref{eq:specular_shading}
 and~\eqref{eq:specular_Albedo minimization}, where the
 IR lighting direction $\vec{l}_p$ is calculated using
 the camera-projector calibration parameters. 
In the natural lighting scene we use the relevant
 normalized coefficients of the first and second order
 spherical harmonics in order to compute the general
 lighting direction. 
From the results in Table~\ref{tab:specular_results} we
 can infer that the specular irradiance can be
 accurately estimated in our proposed lighting model as
 opposed to the natural lighting (NL SH1/2) where
 estimation errors are much larger. 
The reason for large differences is that, as opposed to
 our lighting model, under natural illumination there
 are usually multiple light sources that cause
 specularities whose directions cannot be recovered accurately.
An example of this can be seen in
 Figure~\ref{fig:specular_test}. 
\begin{figure}
\begin{subfigure}{0.14\textwidth}
\includegraphics[trim = 20mm 20mm 20mm 20mm, clip=true, height = 3cm]{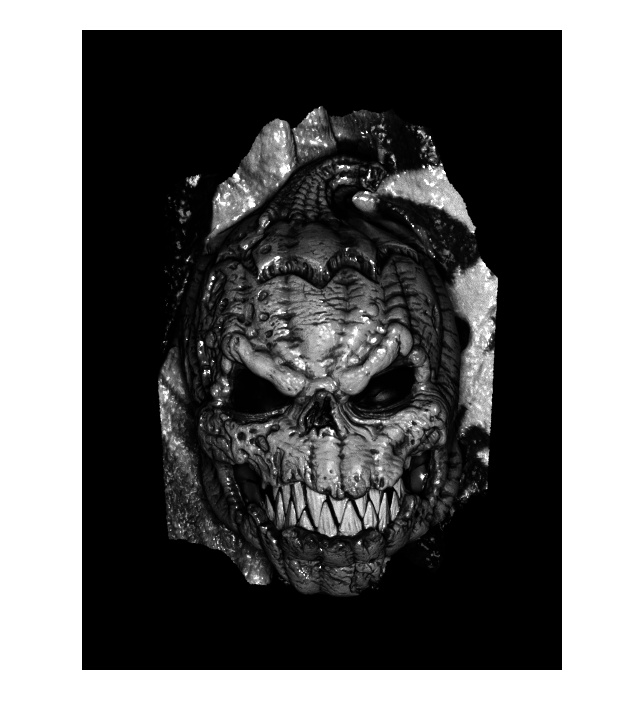}
\caption{}
\end{subfigure}
\hspace{0.01\textwidth}
\begin{subfigure}{0.14\textwidth}
\includegraphics[trim = 90mm 30mm 85mm 10mm, clip=true, height = 2.8cm]{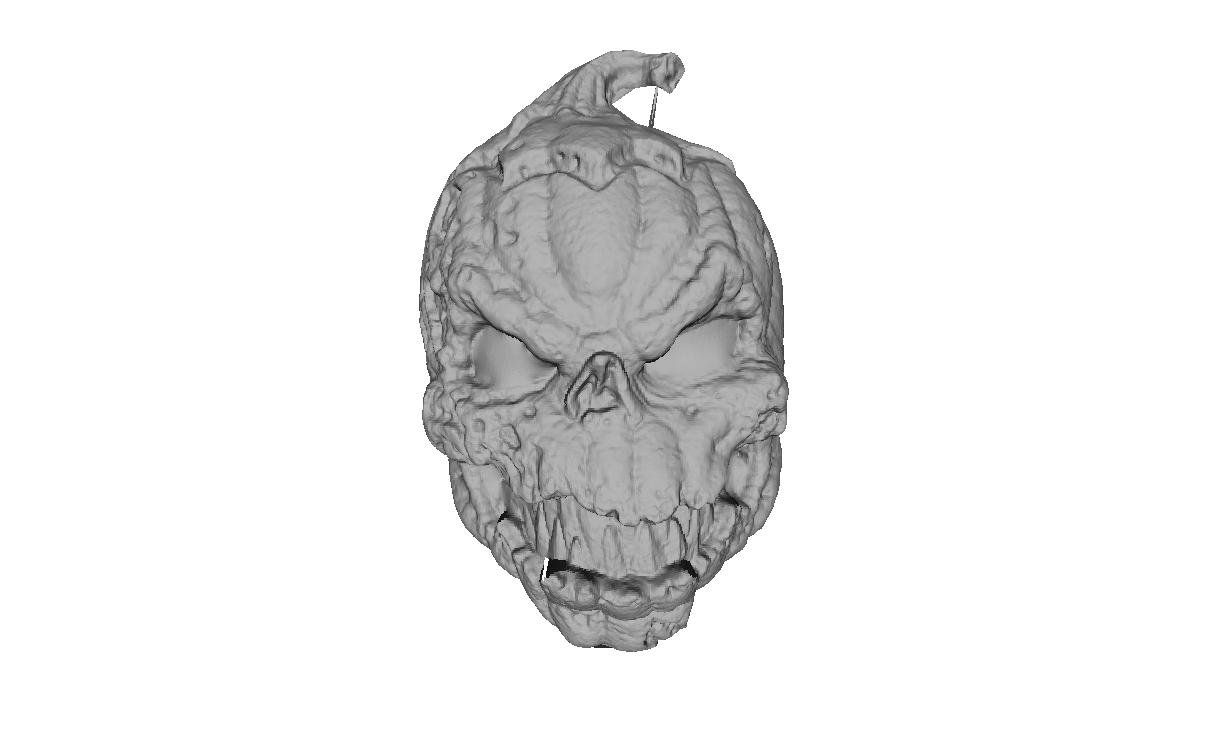}
\caption{}
\end{subfigure}
\hspace{0.01\textwidth}
\begin{subfigure}{0.14\textwidth}
\includegraphics[trim = 90mm 30mm 85mm 10mm, clip=true, height = 2.8cm]{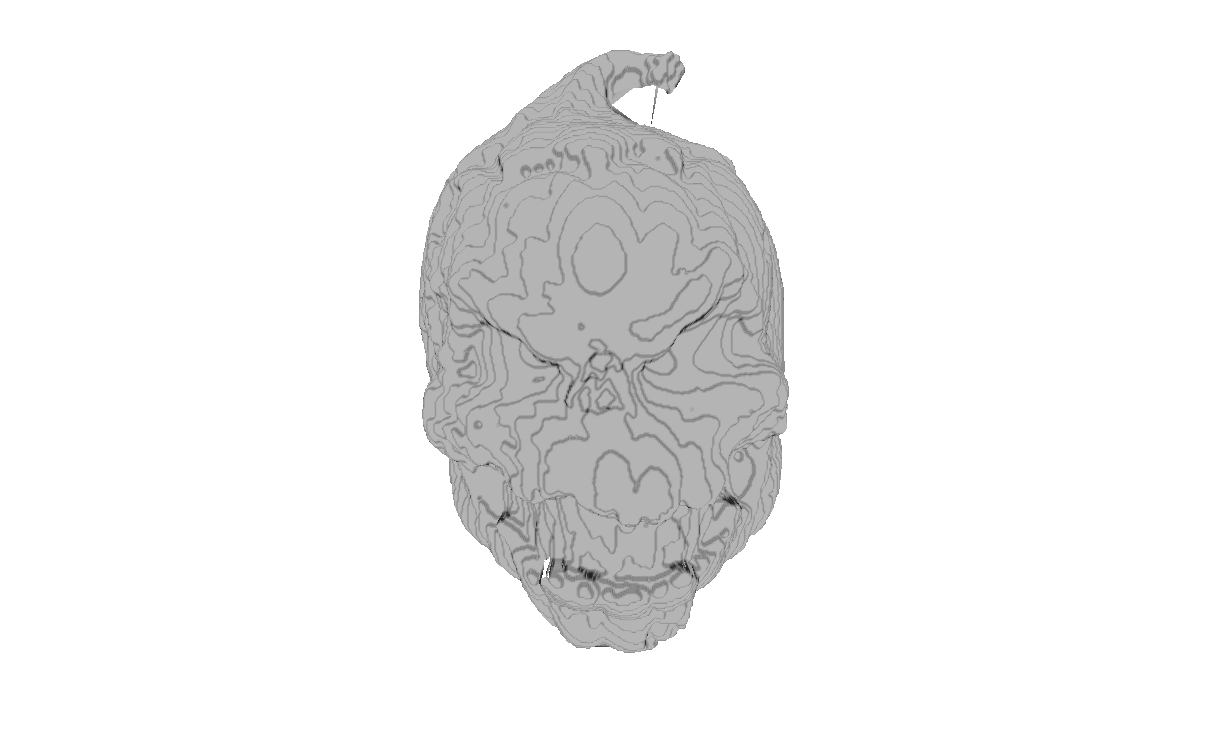}
\caption{}
\end{subfigure}\\
\begin{subfigure}{0.14\textwidth}
\begin{tikzpicture}
\node [anchor=south west,inner sep=0] (image) at (0,0) {\includegraphics[trim = 90mm 30mm 85mm 10mm, clip=true, height = 2.8cm]{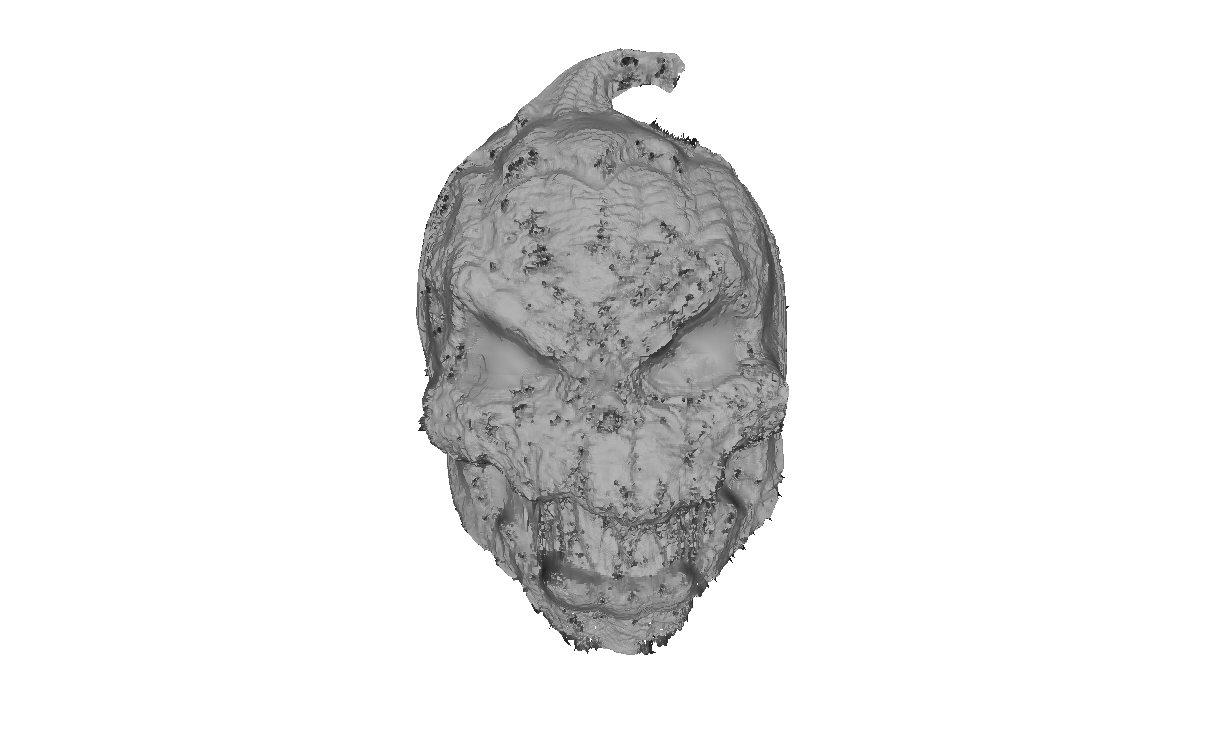}};
\begin{scope}[x={(image.south east)},y={(image.north west)}]
\draw[black,very thick] (0.42,0.42) rectangle (0.62,0.62);
\end{scope}
\end{tikzpicture}
\caption{}
\end{subfigure}
\hspace{0.02\textwidth}
\begin{subfigure}{0.14\textwidth}
\begin{tikzpicture}
\node [anchor=south west,inner sep=0] (image) at (0,0) {\includegraphics[trim = 90mm 30mm 85mm 10mm, clip=true, height = 2.8cm]{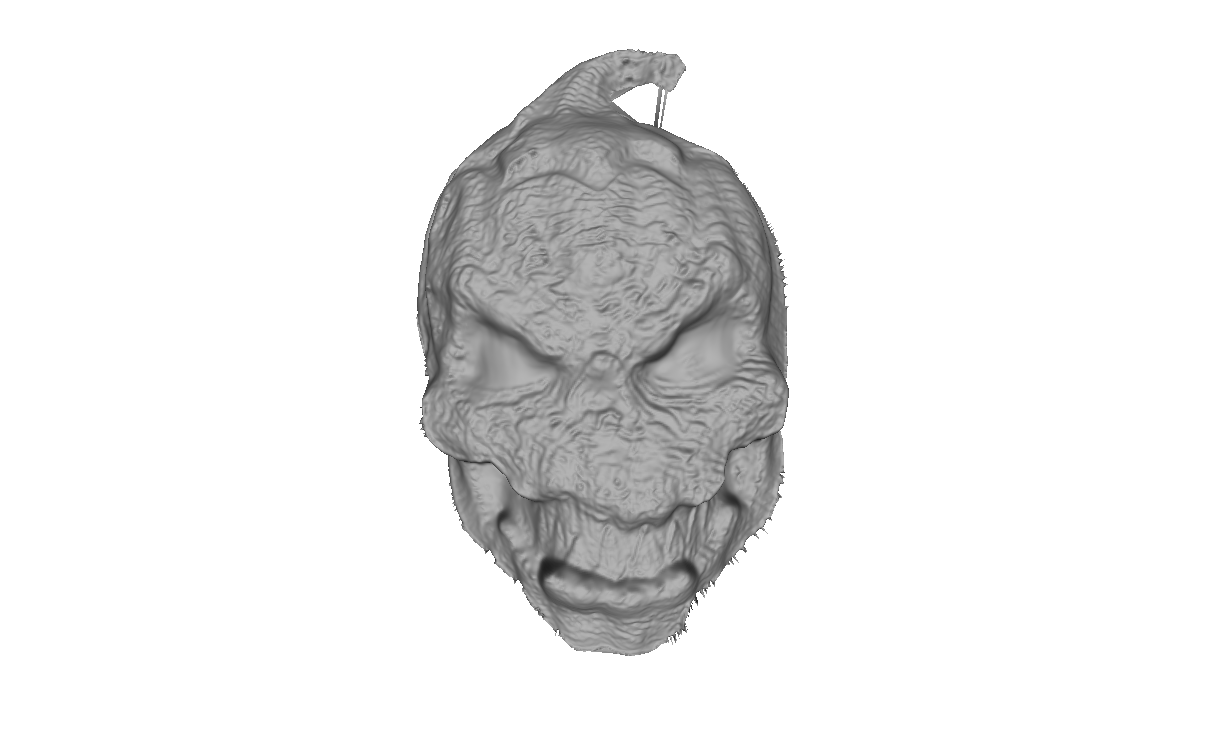}};
\begin{scope}[x={(image.south east)},y={(image.north west)}]
\draw[red,very thick] (0.42,0.42) rectangle (0.62,0.62);
\end{scope}
\end{tikzpicture}
\caption{}
\end{subfigure}
\hspace{0.02\textwidth}
\begin{subfigure}{0.14\textwidth}
\begin{tikzpicture}
\node [anchor=south west,inner sep=0] (image) at (0,0) {\includegraphics[trim = 90mm 30mm 85mm 10mm, clip=true, height = 2.8cm]{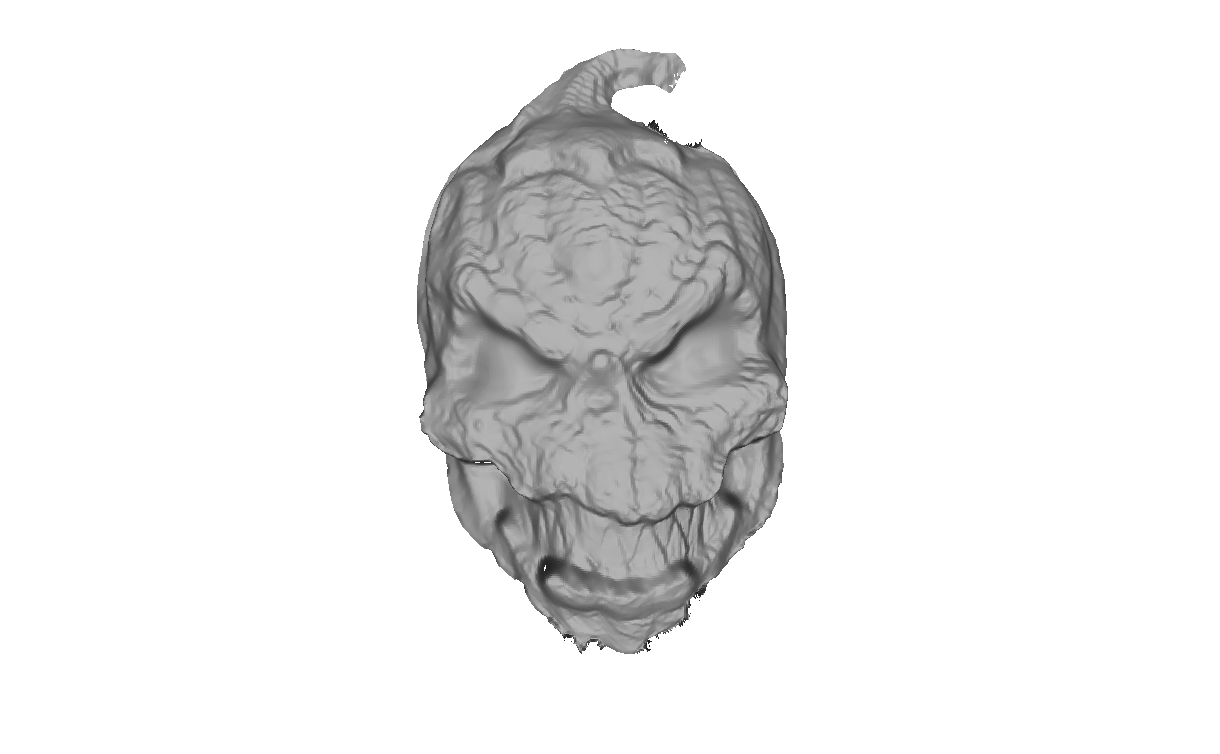}};
\begin{scope}[x={(image.south east)},y={(image.north west)}]
\draw[blue,very thick] (0.42,0.42) rectangle (0.62,0.62);
\end{scope}
\end{tikzpicture}
\caption{}
\end{subfigure}\\
\begin{subfigure}{0.14\textwidth}
\begin{tikzpicture}
\node[anchor=south west,inner sep=0] (image) at (0,0) {\includegraphics[trim = 150mm 97mm 140mm 67mm, clip=true, height = 2.7cm]{graphics/pumpkin_wu00.png}};
\begin{scope}[x={(image.south west)},y={(image.north east)}]
\draw[black,very thick] (0.03,0) rectangle (0.97,1);
\end{scope}
\end{tikzpicture}
\caption{}
\end{subfigure}
\hspace{0.01\textwidth}
\begin{subfigure}{0.14\textwidth}
\begin{tikzpicture}
\node[anchor=south west,inner sep=0] (image) at (0,0) {\includegraphics[trim = 150mm 97mm 140mm 67mm, clip=true, height = 2.7cm]{graphics/pumpkin_rgbd_fusion00.png}};
\begin{scope}[x={(image.south west)},y={(image.north east)}]
\draw[red,very thick] (0.03,0) rectangle (0.97,1);
\end{scope}
\end{tikzpicture}
\caption{}
\end{subfigure}
\hspace{0.01\textwidth}
\begin{subfigure}{0.14\textwidth}
\begin{tikzpicture}
\node[anchor=south west,inner sep=0] (image) at (0,0) {\includegraphics[trim = 150mm 97mm 140mm 67mm, clip=true, height = 2.7cm]{graphics/pumpkin_ours00.png}};
\begin{scope}[x={(image.south west)},y={(image.north east)}]
\draw[blue,very thick] (0.03,0) rectangle (0.97,1);
\end{scope}
\end{tikzpicture}
\caption{}
\end{subfigure}
\vspace{-2mm}
\caption{Results for the simulated Pumpkin scene, (a) Input IR image. (b) Ground truth model. (c) Initial Depth. (d)-(f) Reconstructions of Wu \etal, Or - El \etal and our proposed method respectively. (g)-(i) Magnifications of a specular area. Note the lack of hallucinated features in our method.}
\label{fig:pumpkin_results}
\vspace{-6mm}
\end{figure}

To measure the depth reconstruction accuracy of the
 proposed method we performed experiments using both
 synthetic and real data. 
In the first experiment, we used the $3D$ models with
 mixed diffuse and specular shaders and rendered their
 IR image and ground truth depth maps in Blender. 
We then quantized the ground truth depth map to $1.5$mm
 units in order to simulate the noise of a depth sensor.
We applied our method to the data and defined the
 reconstruction error as the absolute difference between
 the result and the ground truth depth maps. 
We compared our method's performance with the methods
 proposed in~\cite{Orel2015CVPR,WZNSIT14}. 
The comparisons were performed in the specular regions
 of the objects according to the ground truth
 specularity maps. 
The results are shown in
 Table.~\ref{tab:synthetic_results}. 
A qualitative evaluation of the accuracy when the
 method is applied to the synthetic data can be seen
 in  Figures.~\ref{fig:armadillo_results} and
 ~\ref{fig:pumpkin_results}.

In the second experiment we tested our method under 
 laboratory conditions using a structured-light $3D$ 
 scanner 
 to capture the depth of several objects. 
The camera-projector system was calibrated according to
 the method suggested in~\cite{zhang2006novel}. 
We reduced the number of projected patterns in order to
 obtain a noisy depth profile. 
To approximate an IR lighting scenario, we used a
 monochromatic projector and camera with dim ambient
 illumination.   
\begin{figure}
\begin{subfigure}{0.14\textwidth}
\includegraphics[trim = 20mm 20mm 20mm 20mm, clip=true, height = 3cm]{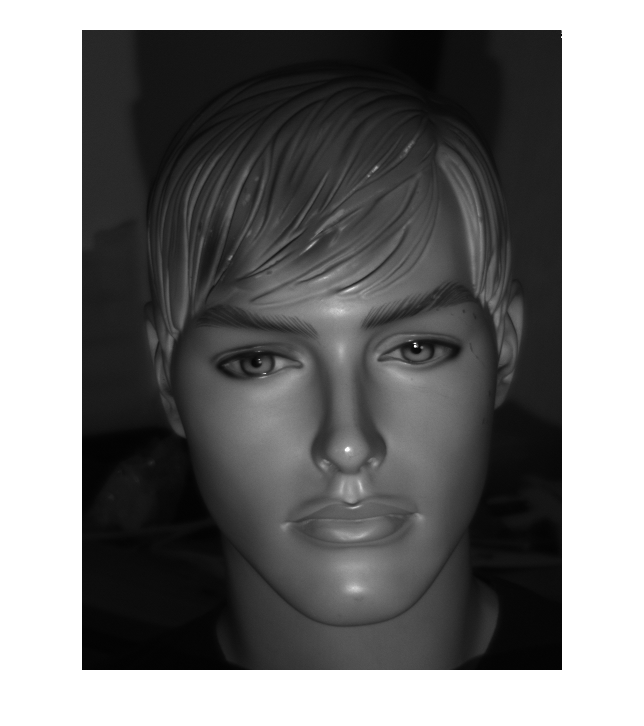}
\caption{}
\end{subfigure}
\hspace{0.01\textwidth}
\begin{subfigure}{0.15\textwidth}
\includegraphics[trim = 90mm 30mm 85mm 10mm, clip=true, height = 2.8cm]{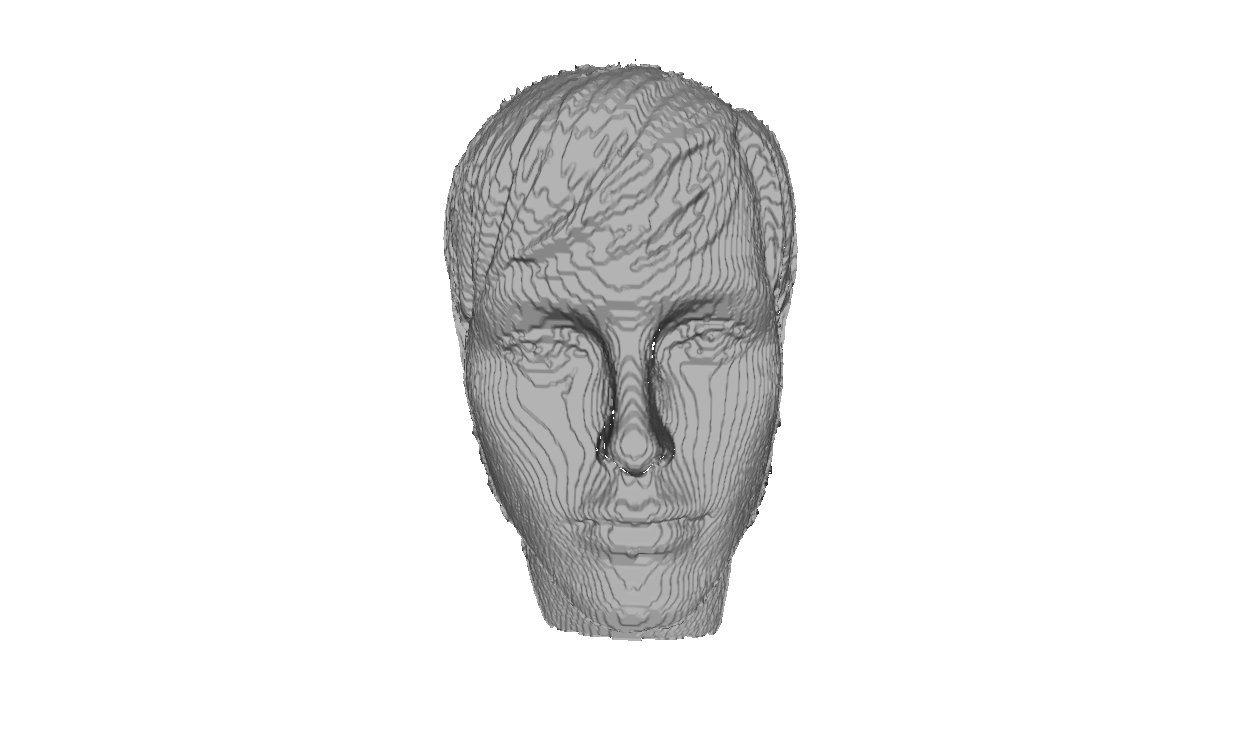}
\caption{}
\end{subfigure}
\hspace{0.01\textwidth}
\begin{subfigure}{0.15\textwidth}
\includegraphics[trim = 90mm 30mm 85mm 10mm, clip=true, height = 2.8cm]{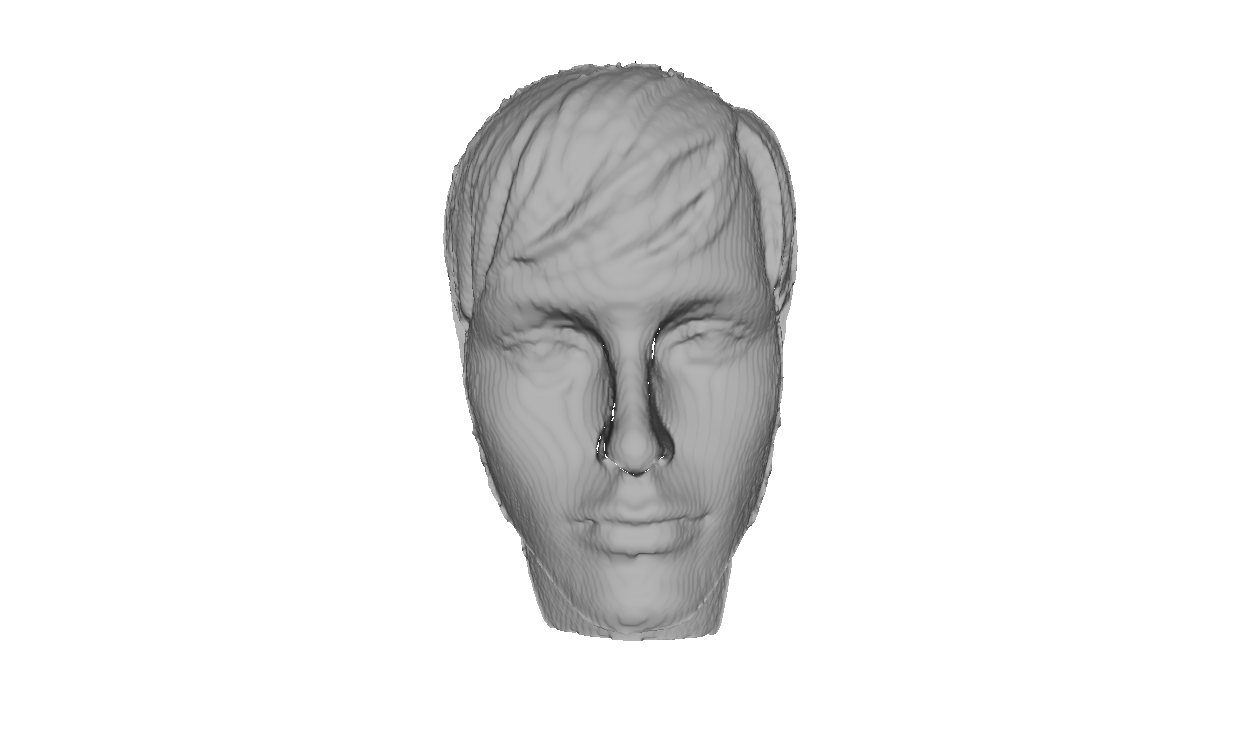}
\caption{}
\end{subfigure}\\
\begin{subfigure}{0.15\textwidth}
\begin{tikzpicture}
\node [anchor=south west,inner sep=0] (image) at (0,0) {\includegraphics[trim = 90mm 30mm 85mm 10mm, clip=true, height = 2.8cm]{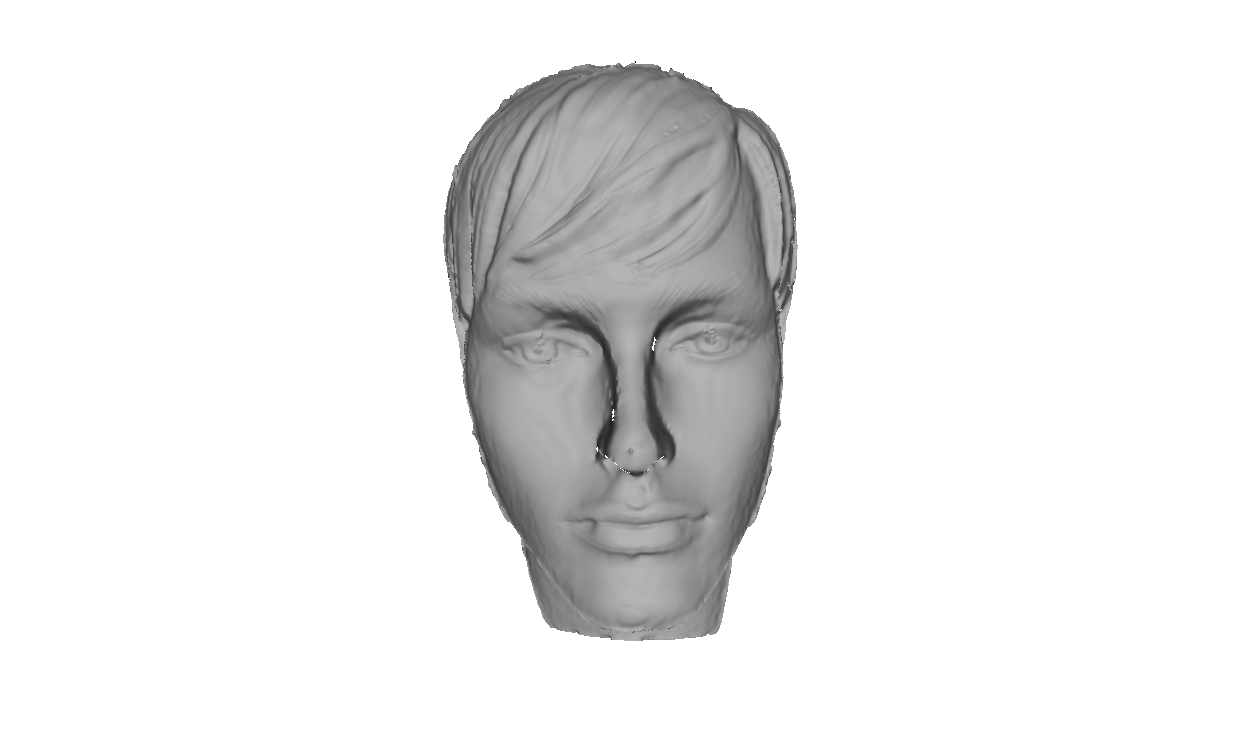}};
\begin{scope}[x={(image.south east)},y={(image.north west)}]
\draw[black,very thick] (0.40,0.20) rectangle (0.60,0.40);
\end{scope}
\end{tikzpicture}
\caption{}
\end{subfigure}
\hspace{0.005\textwidth}
\begin{subfigure}{0.15\textwidth}
\begin{tikzpicture}
\node [anchor=south west,inner sep=0] (image) at (0,0) {\includegraphics[trim = 90mm 30mm 85mm 10mm, clip=true, height = 2.8cm]{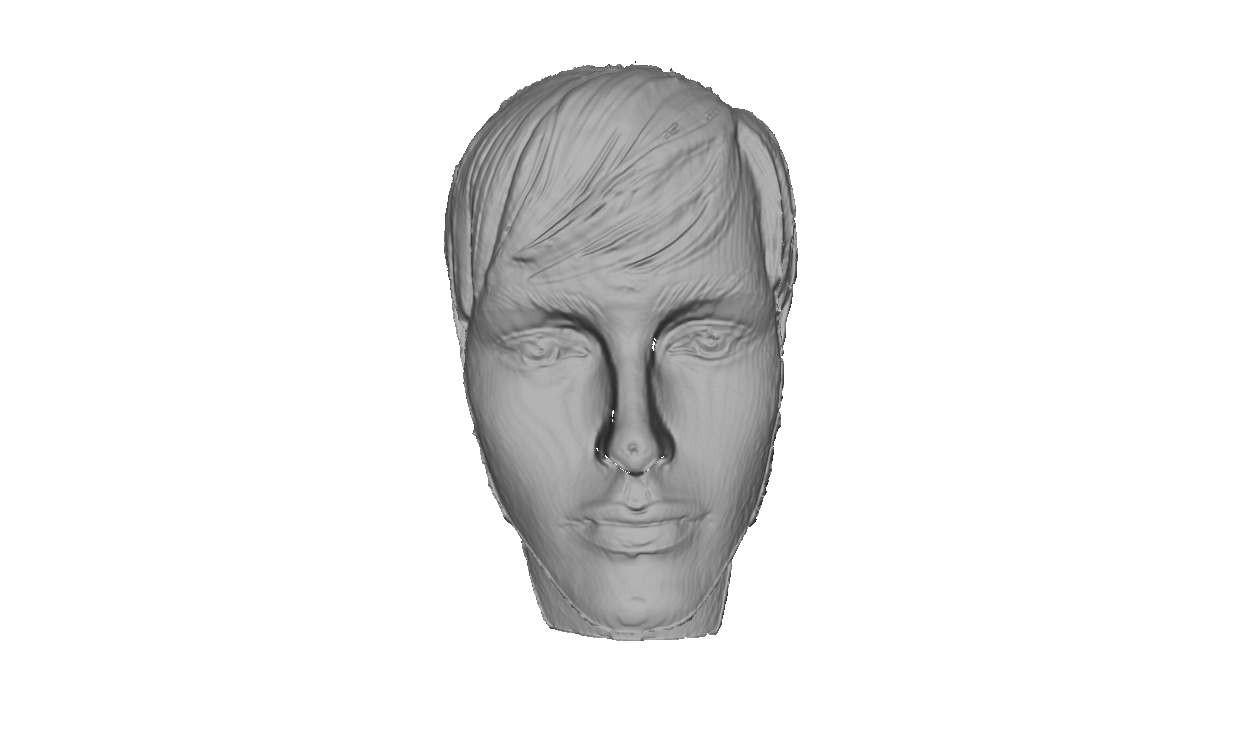}};
\begin{scope}[x={(image.south east)},y={(image.north west)}]
\draw[red,very thick] (0.40,0.20) rectangle (0.60,0.40);
\end{scope}
\end{tikzpicture}
\caption{}
\end{subfigure}
\hspace{0.005\textwidth}
\begin{subfigure}{0.15\textwidth}
\begin{tikzpicture}
\node [anchor=south west,inner sep=0] (image) at (0,0) {\includegraphics[trim = 90mm 30mm 85mm 10mm, clip=true, height = 2.8cm]{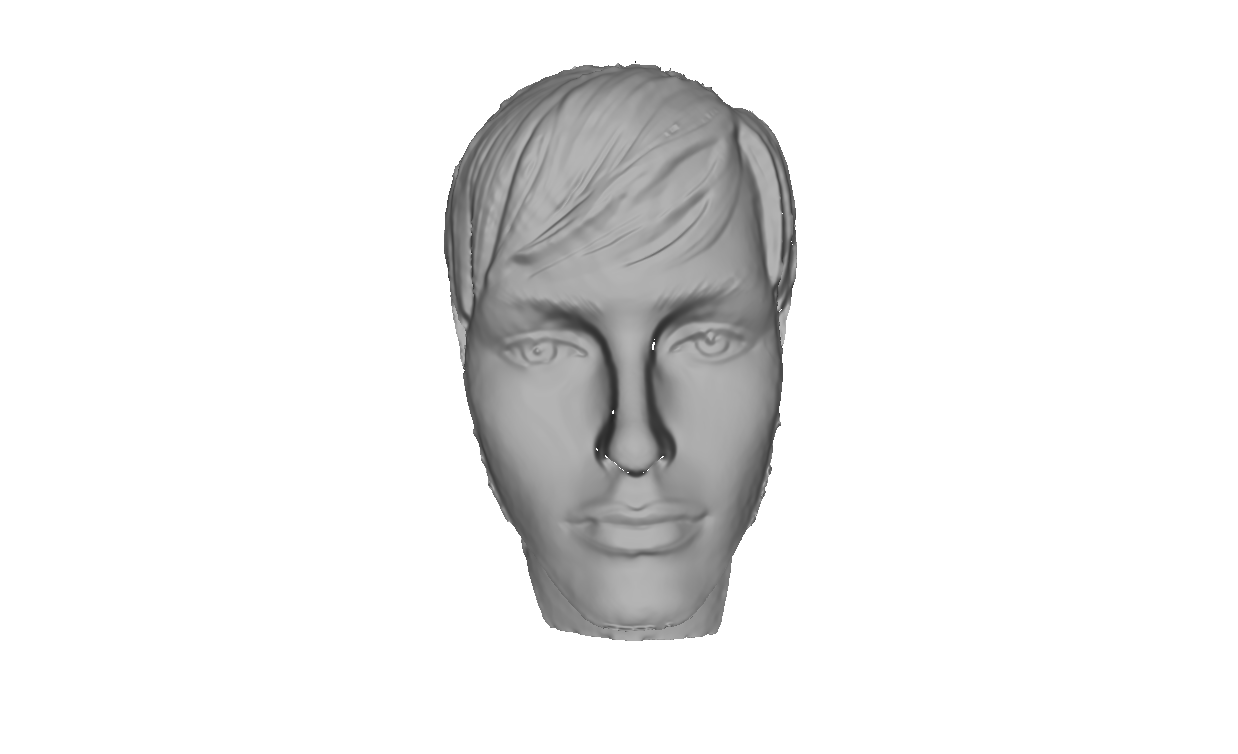}};
\begin{scope}[x={(image.south east)},y={(image.north west)}]
\draw[blue,very thick] (0.40,0.20) rectangle (0.60,0.40);
\end{scope}
\end{tikzpicture}
\caption{}
\end{subfigure}\\
\begin{subfigure}{0.14\textwidth}
\begin{tikzpicture}
\node[anchor=south west,inner sep=0] (image) at (0,0) {\includegraphics[trim = 154mm 60mm 146mm 110mm, clip=true, height = 2.7cm]{graphics/adam_wu00.png}};
\begin{scope}[x={(image.south west)},y={(image.north east)}]
\draw[black,very thick] (0.03,0) rectangle (0.97,1);
\end{scope}
\end{tikzpicture}
\caption{}
\end{subfigure}
\hspace{0.02\textwidth}
\begin{subfigure}{0.14\textwidth}
\begin{tikzpicture}
\node[anchor=south west,inner sep=0] (image) at (0,0) {\includegraphics[trim = 154mm 60mm 146mm 110mm, clip=true, height = 2.7cm]{graphics/adam_rgbd_fusion00.png}};
\begin{scope}[x={(image.south west)},y={(image.north east)}]
\draw[red,very thick] (0.03,0) rectangle (0.97,1);
\end{scope}
\end{tikzpicture}
\caption{}
\end{subfigure}
\hspace{0.02\textwidth}
\begin{subfigure}{0.14\textwidth}
\begin{tikzpicture}
\node[anchor=south west,inner sep=0] (image) at (0,0) {\includegraphics[trim = 154mm 60mm 146mm 110mm, clip=true, height = 2.7cm]{graphics/adam_ours00.png}};
\begin{scope}[x={(image.south west)},y={(image.north east)}]
\draw[blue,very thick] (0.03,0) rectangle (0.97,1);
\end{scope}
\end{tikzpicture}
\caption{}
\end{subfigure}
\vspace{-2mm}
\caption{Results for the lab conditions experiment, 
 (a) Input IR image. (b) Initial Depth. 
 (c) Result after bilateral smoothing. 
 (d)-(f) Reconstructions of Wu \etal, Or - El \etal 
 and the proposed method, respectively. 
 (g)-(i) Magnifications of a specular region.}
\label{fig:adam_results}
\vspace{-5mm}
\end{figure}

We also tested the algorithm with an Intel Real-Sense
 depth scanner, using the IR image and depth map as
 inputs. 
The camera-projector calibration parameters were
 acquired from the Real-Sense SDK platform.    
\begin{figure}
\begin{subfigure}{0.14\textwidth}
\includegraphics[trim = 70mm 15mm 40mm 15mm, clip=true, height = 3cm]{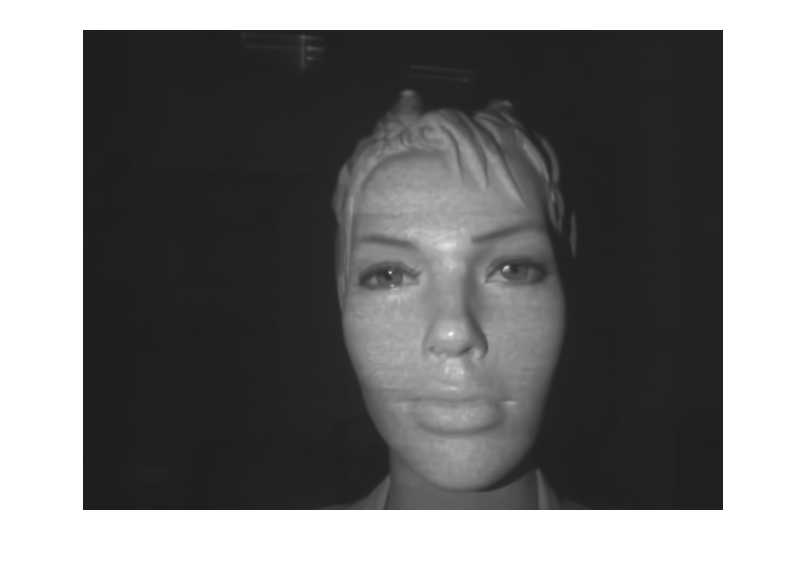}
\caption{}
\end{subfigure}
\hspace{0.01\textwidth}
\begin{subfigure}{0.15\textwidth}
\includegraphics[trim = 90mm 37mm 100mm 0mm, clip=true, height = 2.8cm]{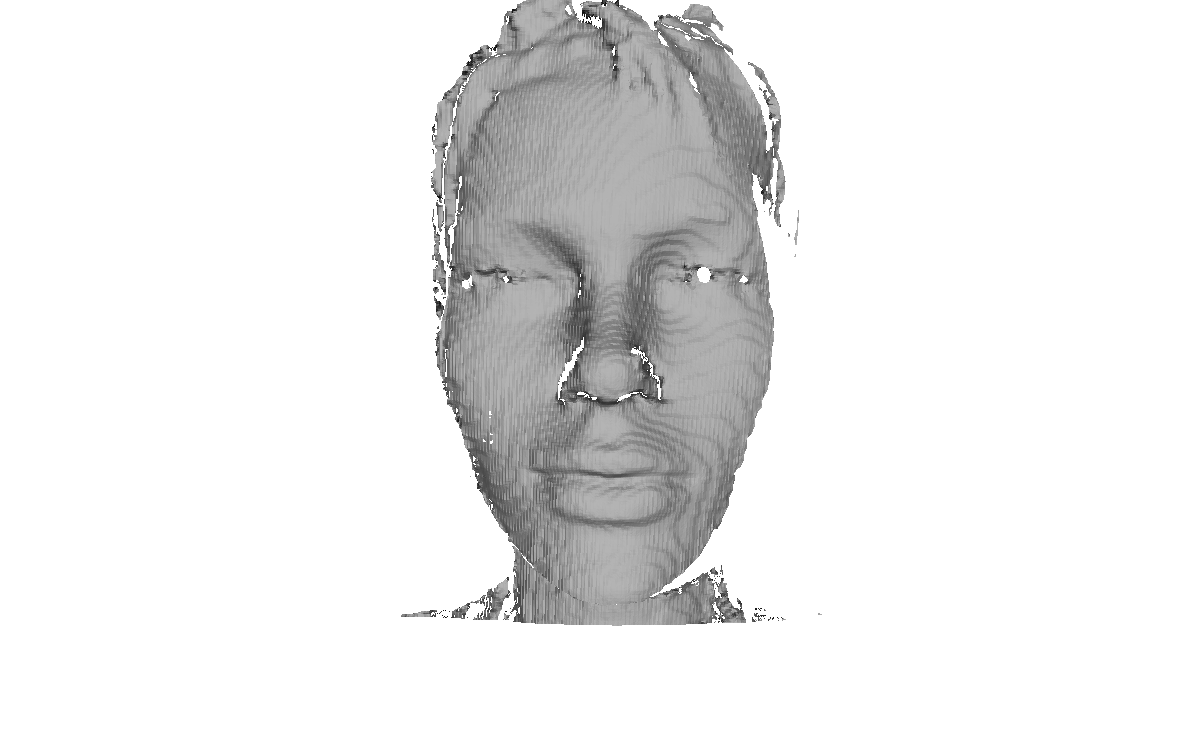}
\caption{}
\end{subfigure}
\hspace{0.01\textwidth}
\begin{subfigure}{0.15\textwidth}
\includegraphics[trim = 90mm 37mm 100mm 0mm, clip=true, height = 2.8cm]{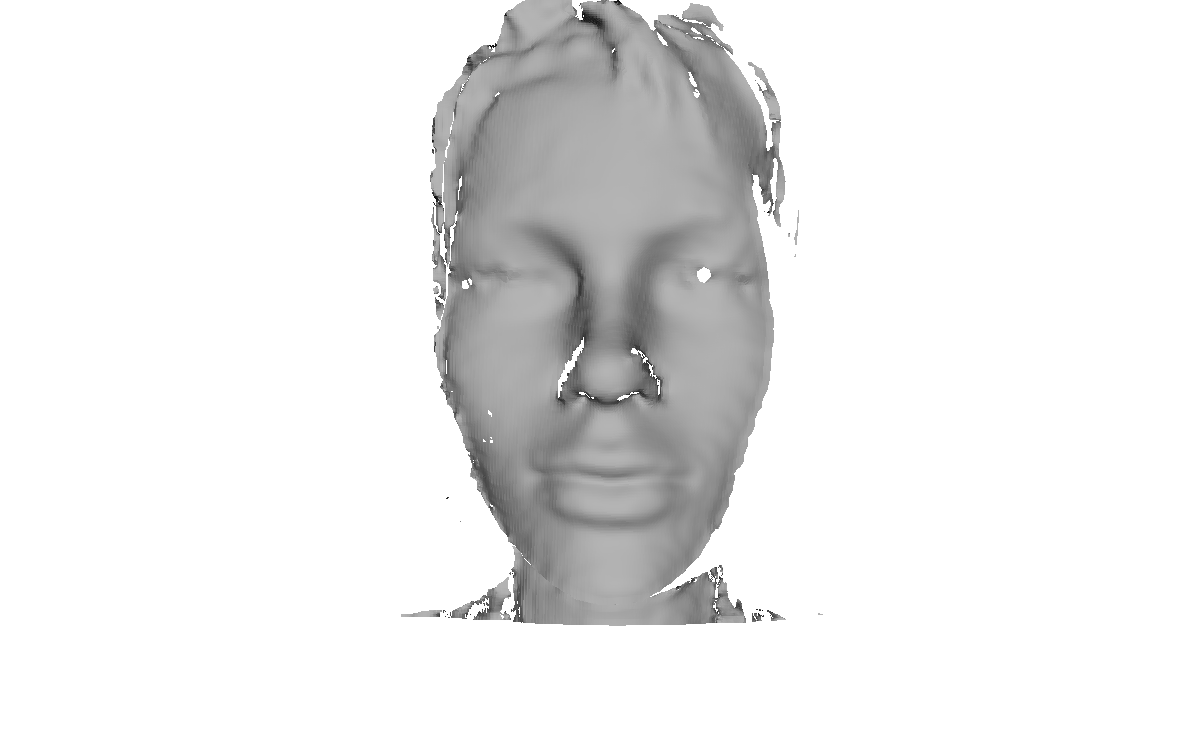}
\caption{}
\end{subfigure}\\
\begin{subfigure}{0.15\textwidth}
\begin{tikzpicture}
\node [anchor=south west,inner sep=0] (image) at (0,0) {\includegraphics[trim = 90mm 37mm 100mm 0mm, clip=true, height = 2.8cm]{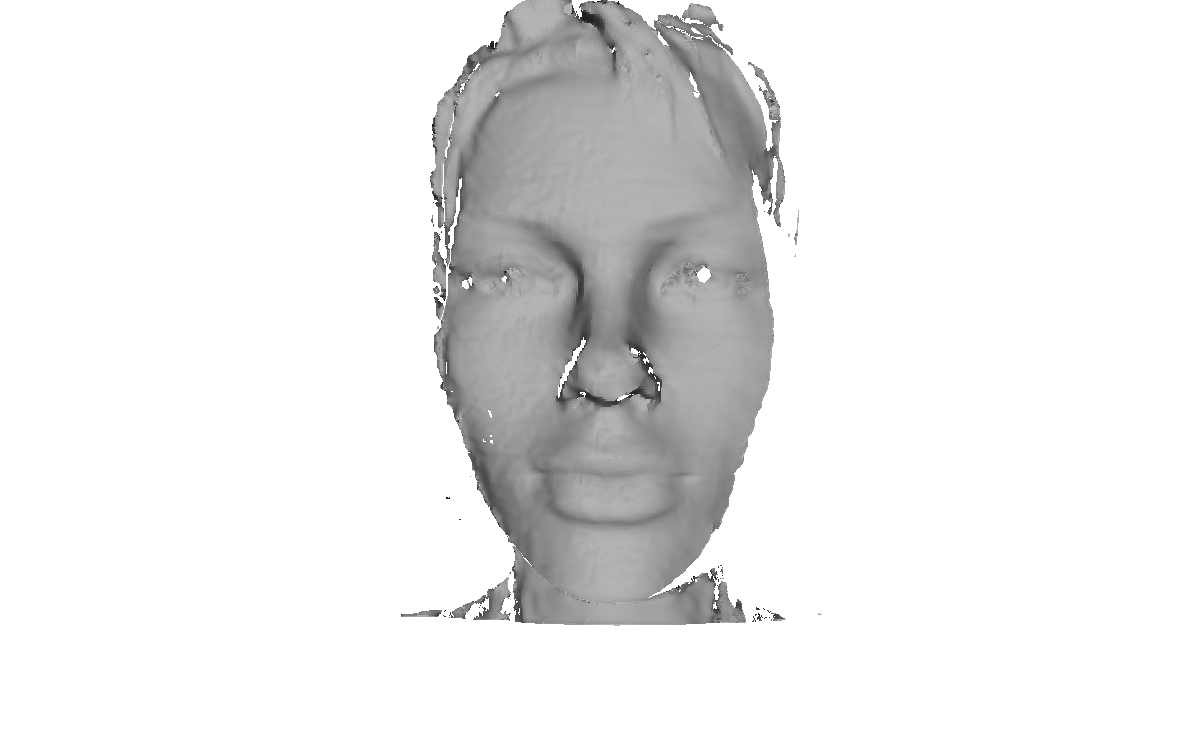}};
\begin{scope}[x={(image.south east)},y={(image.north west)}]
\draw[black,very thick] (0.45,0.30) rectangle (0.65,0.50);
\end{scope}
\end{tikzpicture}
\caption{}
\end{subfigure}
\hspace{0.005\textwidth}
\begin{subfigure}{0.15\textwidth}
\begin{tikzpicture}
\node [anchor=south west,inner sep=0] (image) at (0,0) {\includegraphics[trim = 90mm 37mm 100mm 0mm, clip=true, height = 2.8cm]{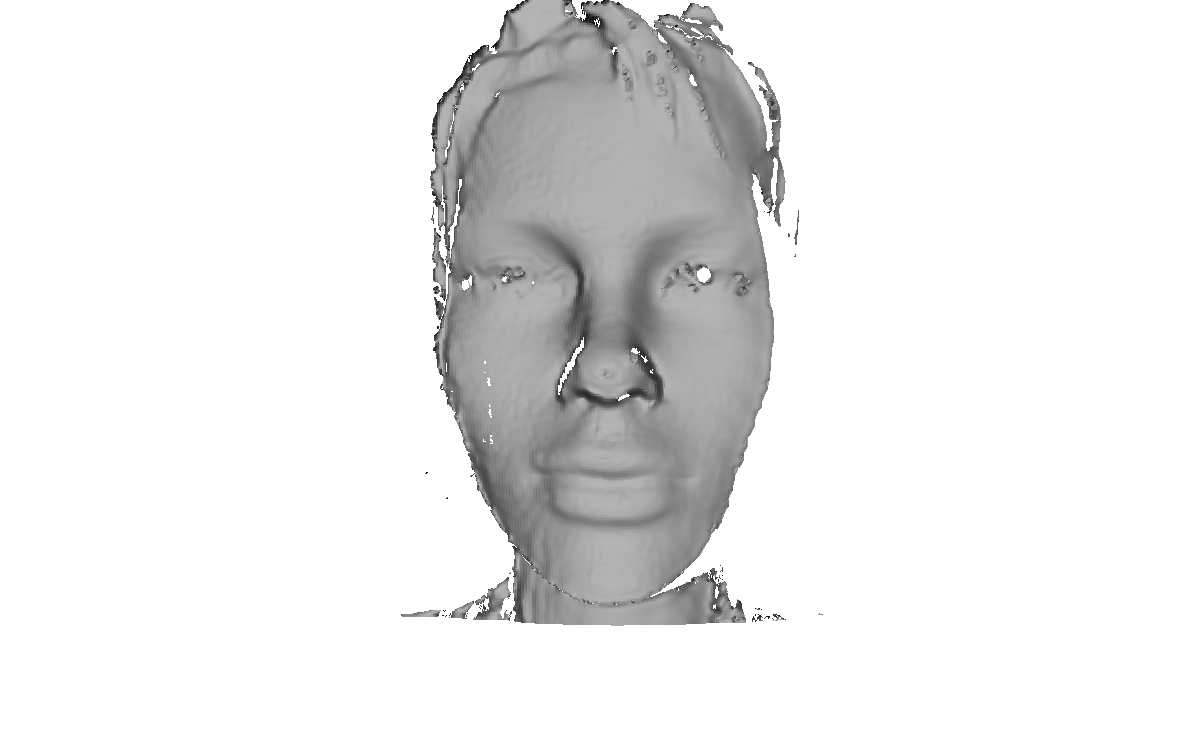}};
\begin{scope}[x={(image.south east)},y={(image.north west)}]
\draw[red,very thick] (0.45,0.30) rectangle (0.65,0.50);
\end{scope}
\end{tikzpicture}
\caption{}
\end{subfigure}
\hspace{0.005\textwidth}
\begin{subfigure}{0.15\textwidth}
\begin{tikzpicture}
\node [anchor=south west,inner sep=0] (image) at (0,0) {\includegraphics[trim = 90mm 37mm 100mm 0mm, clip=true, height = 2.8cm]{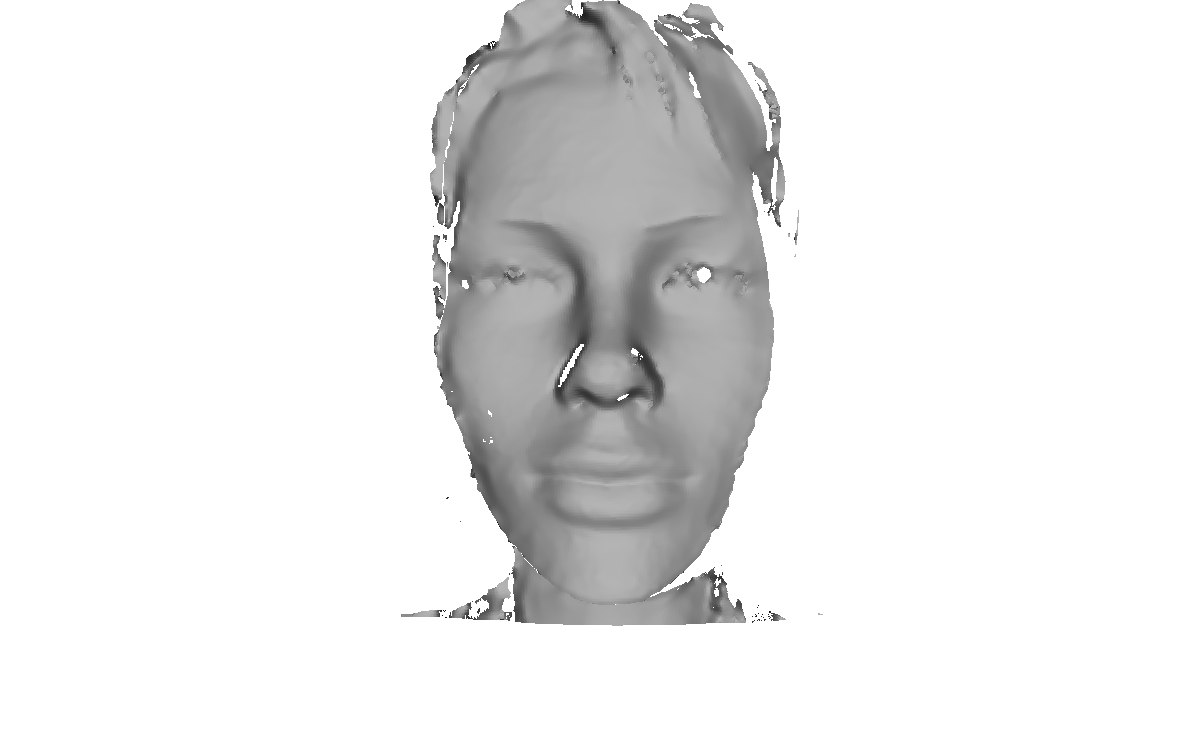}};
\begin{scope}[x={(image.south east)},y={(image.north west)}]
\draw[blue,very thick] (0.45,0.30) rectangle (0.65,0.50);
\end{scope}
\end{tikzpicture}
\caption{}
\end{subfigure}\\
\begin{subfigure}{0.14\textwidth}
\begin{tikzpicture}
\node[anchor=south west,inner sep=0] (image) at (0,0) {\includegraphics[trim = 147mm 90mm 143mm 82mm, clip=true, height = 2.5cm]{graphics/liz_realsense_wu00.png}};
\begin{scope}[x={(image.south west)},y={(image.north east)}]
\draw[black,very thick] (0.03,0) rectangle (0.97,1);
\end{scope}
\end{tikzpicture}
\caption{}
\end{subfigure}
\hspace{0.015\textwidth}
\begin{subfigure}{0.14\textwidth}
\begin{tikzpicture}
\node[anchor=south west,inner sep=0] (image) at (0,0) {\includegraphics[trim = 147mm 90mm 143mm 82mm, clip=true, height = 2.5cm]{graphics/liz_realsense_rgbd_fusion00.png}};
\begin{scope}[x={(image.south west)},y={(image.north east)}]
\draw[red,very thick] (0.03,0) rectangle (0.97,1);
\end{scope}
\end{tikzpicture}
\caption{}
\end{subfigure}
\hspace{0.015\textwidth}
\begin{subfigure}{0.14\textwidth}
\begin{tikzpicture}
\node[anchor=south west,inner sep=0] (image) at (0,0) {\includegraphics[trim = 147mm 90mm 143mm 82mm, clip=true, height = 2.5cm]{graphics/liz_realsense_ours00.png}};
\begin{scope}[x={(image.south west)},y={(image.north east)}]
\draw[blue,very thick] (0.03,0) rectangle (0.97,1);
\end{scope}
\end{tikzpicture}
\caption{}
\end{subfigure}
\caption{Results from Intel's Real-Sense depth scanner,
 (a) Input IR image. 
 (b) Initial Depth. 
 (c) Result after bilateral smoothing. 
 (d)-(f) Reconstructions of Wu \etal, Or - El \etal and
 the proposed method, respectively. 
 (g)-(i) Magnifications of a specular region.}
\label{fig:liz_results}
\vspace{-6mm}
\end{figure}
Although no accurate ground-truth data was available for
 these experiments, we note that while all methods
 exhibit sufficient accuracy in diffuse areas, the 
 proposed method is the only one that performs
 qualitatively well in highly specular areas as can be
 seen in Figures~\ref{fig:adam_results} 
 and~\ref{fig:liz_results}. 

\section{Conclusions}
\label{sec:Conclusions}
We presented a new framework for depth refinement of
 specular objects based on shading cues from an IR
 image.
To the best of our knowledge, the proposed method is
 the first depth refinement framework to explicitly
 account for specular lighting. 
An efficient optimization scheme enables our system to
 produce state of the art results at real-time rates.  

\section*{Acknowledgments}
We wish to thank Alon Zvirin for his help with the experiments. 
This research was supported by European Community’s FP7-ERC program
grant agreement no. 267414. G.R is partially funded by VITALITE Army Research Office Multidisciplinary Research Initiative program, award W911NF-11-1-0391. 

{\small
\bibliographystyle{ieee}
\bibliography{mybib}

\begin{thebibliography}{10}\itemsep=-1pt

\bibitem{BarronTPAMI2015}
J.~T. Barron and J.~Malik.
\newblock Shape, illumination, and reflectance from shading.
\newblock {\em IEEE Transactions on Pattern Analysis and Machine Intelligence},
  pages 1670--1687, 2015.

\bibitem{basri2003lambertian}
R.~Basri and D.~W. Jacobs.
\newblock Lambertian reflectance and linear subspaces.
\newblock {\em IEEE Transactions on Pattern Analysis and Machine Intelligence},
  25(2):218--233, 2003.

\bibitem{bohme2010shading}
M.~B{\"o}hme, M.~Haker, T.~Martinetz, and E.~Barth.
\newblock Shading constraint improves accuracy of time-of-flight measurements.
\newblock {\em Computer vision and image understanding}, 114(12):1329--1335,
  2010.

\bibitem{Chatterjee2015CVPR}
A.~Chatterjee and V.~M. Govindu.
\newblock Photometric refinement of depth maps for multi-albedo objects.
\newblock In {\em IEEE Conference on Computer Vision and Pattern Recognition
  (CVPR)}, pages 933--941, 2015.

\bibitem{Choe2014CVPR}
G.~Choe, J.~Park, Y.-W. Tai, and I.~So~Kweon.
\newblock Exploiting shading cues in kinect {IR} images for geometry
  refinement.
\newblock In {\em IEEE Conference on Computer Vision and Pattern Recognition
  (CVPR)}, pages 3922--3929, 2014.

\bibitem{durou2008numerical}
J.-D. Durou, M.~Falcone, and M.~Sagona.
\newblock Numerical methods for shape-from-shading: A new survey with
  benchmarks.
\newblock {\em Computer Vision and Image Understanding}, 109(1):22--43, 2008.

\bibitem{hanhigh2013}
Y.~Han, J.~Y. Lee, and I.~S. Kweon.
\newblock High quality shape from a single {RGB-D} image under uncalibrated
  natural illumination.
\newblock In {\em IEEE International Conference on Computer Vision (ICCV)},
  pages 1617--1624, 2013.

\bibitem{haque2014high}
S.~Haque, A.~Chatterjee, and V.~M. Govindu.
\newblock High quality photometric reconstruction using a depth camera.
\newblock In {\em IEEE Conference on Computer Vision and Pattern Recognition
  (CVPR)}, pages 2283--2290, 2014.

\bibitem{kadambi2015polarized}
A.~Kadambi, V.~Taamazyan, B.~Shi, and R.~Raskar.
\newblock Polarized 3{D}: High-quality depth sensing with polarization cues.
\newblock In {\em IEEE International Conference on Computer Vision}, pages
  3370--3378, 2015.

\bibitem{categoryShapesKar15}
A.~Kar, S.~Tulsiani, J.~Carreira, and J.~Malik.
\newblock Category-specific object reconstruction from a single image.
\newblock In {\em IEEE Conference on Computer Vision and Pattern Recognition
  (CVPR)}, pages 1966--1974. 2015.

\bibitem{newcombe2011kinectfusion}
R.~A. Newcombe, A.~J. Davison, S.~Izadi, P.~Kohli, O.~Hilliges, J.~Shotton,
  D.~Molyneaux, S.~Hodges, D.~Kim, and A.~Fitzgibbon.
\newblock Kinect{F}usion: Real-time dense surface mapping and tracking.
\newblock In {\em IEEE international symposium on Mixed and augmented reality},
  pages 127--136, 2011.

\bibitem{Orel2015CVPR}
R.~Or~El, G.~Rosman, A.~Wetzler, R.~Kimmel, and A.~M. Bruckstein.
\newblock {RGBD-F}usion: Real-time high precision depth recovery.
\newblock In {\em IEEE Conference on Computer Vision and Pattern Recognition
  (CVPR)}, pages 5407--5416, 2015.

\bibitem{ping1994shape}
T.~Ping-Sing and M.~Shah.
\newblock Shape from shading using linear approximation.
\newblock {\em Image and Vision Computing}, 12(8):487--498, 1994.

\bibitem{ramamoorthi2001efficient}
R.~Ramamoorthi and P.~Hanrahan.
\newblock An efficient representation for irradiance environment maps.
\newblock In {\em Proceedings of the 28th annual conference on Computer
  graphics and interactive techniques}, pages 497--500. ACM, 2001.

\bibitem{Richter2015CVPR}
S.~R. Richter and S.~Roth.
\newblock Discriminative shape from shading in uncalibrated illumination.
\newblock In {\em IEEE Conference on Computer Vision and Pattern Recognition
  (CVPR)}, pages 1128--1136, 2015.

\bibitem{rosman2012group}
G.~Rosman, A.~M. Bronstein, M.~M. Bronstein, X.-C. Tai, and R.~Kimmel.
\newblock Group-valued regularization for analysis of articulated motion.
\newblock In {\em NORDIA workshop, European Conference on Computer Vision
  (ECCV)}, pages 52--62. Springer, 2012.

\bibitem{roussos2010tensor}
A.~Roussos and P.~Maragos.
\newblock Tensor-based image diffusions derived from generalizations of the
  total variation and beltrami functionals.
\newblock In {\em IEEE International Conference on Image Processing (ICIP)},
  pages 4141--4144. IEEE, 2010.

\bibitem{sochen1998general}
N.~Sochen, R.~Kimmel, and R.~Malladi.
\newblock A general framework for low level vision.
\newblock {\em IEEE Transactions on Image Processing}, 7(3):310--318, 1998.

\bibitem{Ti_2015_CVPR}
C.~Ti, R.~Yang, J.~Davis, and Z.~Pan.
\newblock Simultaneous time-of-flight sensing and photometric stereo with a
  single tof sensor.
\newblock In {\em IEEE Conference on Computer Vision and Pattern Recognition
  (CVPR)}, pages 4334--4342, 2015.

\bibitem{Wetzler11}
A.~Wetzler and R.~Kimmel.
\newblock Efficient beltrami flow in patch-space.
\newblock In {\em Scale Space and Variational Methods in Computer Vision
  (SSVM)}, pages 134--143, 2011.

\bibitem{Wu:2010}
C.~Wu and X.-C. Tai.
\newblock Augmented lagrangian method, dual methods, and split bregman
  iteration for {ROF}, vectorial {TV}, and high order models.
\newblock {\em SIAM J. Img. Sci.}, 3:300--339, July 2010.

\bibitem{WZNSIT14}
C.~Wu, M.~Zollh{\"o}fer, M.~Nie{\ss}ner, M.~Stamminger, S.~Izadi, and
  C.~Theobalt.
\newblock Real-time shading-based refinement for consumer depth cameras.
\newblock In {\em ACM Transactions on Graphics (Proceedings of SIGGRAPH Asia
  2014)}, volume~33, December 2014.

\bibitem{yu2013shading}
L.~F. Yu, S.~K. Yeung, Y.~W. Tai, and S.~Lin.
\newblock Shading-based shape refinement of {RGB-D} images.
\newblock In {\em IEEE Conference on Computer Vision and Pattern Recognition
  (CVPR)}, 2013.

\bibitem{zhang1999shape}
R.~Zhang, P.-S. Tsai, J.~E. Cryer, and M.~Shah.
\newblock Shape-from-shading: a survey.
\newblock {\em Pattern Analysis and Machine Intelligence, IEEE Transactions
  on}, 21(8):690--706, 1999.

\bibitem{zhang2006novel}
S.~Zhang and P.~S. Huang.
\newblock Novel method for structured light system calibration.
\newblock {\em Optical Engineering}, 45(8):083601--1--083601--8, 2006.

\end{thebibliography}
}

\end{document}